\documentclass[preprint,12pt,authoryear]{elsarticle}

\usepackage{amssymb}
\usepackage{amsthm}
\usepackage{lineno}
\usepackage{soul}

\usepackage{hyperref}
\usepackage[usenames, dvipsnames]{color}

\newcommand{\ola}[1]{\textcolor{black}{#1}} 
\usepackage{siunitx} 
\usepackage{textcomp} 
\usepackage{gensymb} 
\usepackage{booktabs} 
\usepackage{multirow} 
\newcolumntype{C}{>{\centering\arraybackslash}p{3.25em}}

\begin{document}

\begin{frontmatter}

\title{Title\tnoteref{label1}}

\author[label1]{Aleksandra Wolanin}
\corref{cor1}
\ead{ola@gfz-potsdam.de}
\author[label2]{Gustau Camps-Valls}
\ead{Gustau.Camps@uv.es}
\author[label2]{Luis G{\'o}mez-Chova}
\ead{luis.gomez-chova@uv.es}
\author[label2]{Gonzalo Mateo-Garc{\'i}a}
\ead{gonzalo.mateo-garcia@uv.es}
\author[label4]{Christiaan van der Tol}
\ead{c.vandertol@utwente.nl}
\author[label3]{Yongguang Zhang}
\ead{yongguang\_zhang@nju.edu.cn}
\author[label1]{Luis Guanter}
\ead{guanter@gfz-potsdam.de}

\cortext[cor1]{{\bf Preprint of paper published in Remote Sensing of Environment Volume 225, May 2019, Pages 441-457. https://doi.org/10.1016/j.rse.2019.03.002}}

\address[label1]{Section 1.4 Remote Sensing, GFZ German Research Centre for Geosciences, Helmholtz-Centre,  Potsdam,  Germany\fnref{label1}}
\address[label2]{Image Processing Laboratory, Universitat de  Val\`encia, Val\`encia, Spain\fnref{label2}} 
\address[label4]{Faculty of ITC, University of Twente, Enschede, The Netherlands\fnref{label4}}
\address[label3]{Jiangsu Provincial Key Laboratory of Geographic Information Science and Technology, International Institute for Earth System Sciences, Nanjing University, 210023 Nanjing, China;\fnref{label3}}
 
\title{{Estimating Crop Primary Productivity with Sentinel-2 and Landsat 8 using Machine Learning Methods Trained with Radiative Transfer Simulations}}
 
\begin{abstract}
Satellite remote sensing has been widely used in the last decades for agricultural applications, {both for assessing vegetation condition and for subsequent yield prediction.} 
Existing remote sensing-based methods to estimate gross primary productivity (GPP), which is an important variable to indicate crop photosynthetic function and stress,  typically rely on empirical or semi-empirical approaches, which tend to over-simplify  photosynthetic mechanisms.  
In this work, we take advantage of all parallel developments in mechanistic photosynthesis modeling and satellite data availability for an advanced monitoring of crop productivity. In particular, we combine process-based modeling with the soil-canopy energy balance radiative transfer model (SCOPE) with Sentinel-2 {and Landsat 8} optical remote sensing data and machine learning methods in order to estimate crop GPP. With this approach, we by-pass the need for an intermediate step to retrieve the set of vegetation biophysical parameters needed to accurately \ola{model} photosynthesis, while still accounting for the complex processes of the original physically-based model. Several implementations of the machine learning models are tested and validated using simulated and flux tower-based GPP data. Our final neural network model is able to estimate GPP at the tested flux tower sites with $r^2$ of 0.92 and RMSE of \SI{1.38}{gC.d^{-1}.m^{-2}}, which outperforms empirical models based on vegetation indices. {The first test of applicability of this model to Landsat 8 data showed good results ($r^2$ of 0.82 and RMSE of \SI{1.97}{gC.d^{-1}.m^{-2}}), which suggests that our approach can be further applied to other sensors.} Modeling and testing is restricted to C3 crops in this study, but can be extended to C4 crops by producing a new training dataset with SCOPE that accounts for the different photosynthetic pathways. Our model successfully estimates GPP across a variety of C3 crop types and environmental conditions even though it does not use any local information from the corresponding sites. This highlights its potential to map crop productivity from new satellite sensors at a global scale with the help of current Earth observation cloud computing platforms. 
\end{abstract}
 

\end{frontmatter}
 
\section{Introduction}
\label{sec:intro} 
Monitoring spatio-temporal changes in the photosynthetic functioning of agricultural lands is of paramount importance for many societal, environmental and economical challenges within the current scenario of increasing demands of biofuels and food.
{In particular, the accurate estimation of the gross primary productivity (GPP, amount of carbon fixed by plants through photosynthesis) of agricultural lands is key for monitoring, understanding and forecasting crop's status and potential yields. 
GPP at various spatio-temporal scales (field, region, the globe) can be applied in order to compare the impact of different management practices (e.g., tillage or crop rotation) and spatio-temporal variations in geographic and meteorological conditions on crop photosynthesis \citep{falge_seasonality_2002,baker_examining_2005,author_usefulness_2005}.
}   

Remote sensing provides consistent and systematic observations of the Earth surface and has therefore remarkably {contributed to} crop monitoring on large scales. {Satellite observations of crops have been applied for crop vegetation monitoring, crop yield forecasting and management decisions optimization by agriculture companies and sectoral organizations \citep[e.g., ][]{strachan_impact_2002, pinter_remote_2003, mulla_twenty_2013, wu_remote_2014, pulwarty_information_2014}.
} Over the last decade, both {quantity} and quality (including spectral and spatial resolution) of remote sensing data have been steadily increasing \citep{belward_who_2015}. 
For example, the Sentinel-2 mission of the European Copernicus program provides observations at a spatial resolution of 10--20\,m, at multiple spectral bands in visible to shortwave infrared wavelengths  with a 5-day revisit time, a long-term operation commitment and a free and open data policy \citep{drusch2012sentinel}, {which constitutes a great improvement as compared to other previous and current missions in terms of agricultural application.}

GPP is typically modeled with three different approaches:
process-based models (PBMs), semi-empirical light use efficiency (LUE) models \citep[e.g.,][]{zhang_evaluating_2012}, and data-driven statistical models \citep{jung_global_2011,Tramontana_2016}.  
PBMs are based on the mechanistic description of  photosynthetic biochemical processes, usually as described in the Farquhar's photosynthesis model \citep{farquhar_biochemical_1980}. 
GPP is first computed at the leaf level and then scaled-up to the whole canopy. In 
LUE models, GPP is explicitly decoupled into two terms: the amount of absorbed photosynthetically active radiation (APAR) and the LUE, the latter accounting for the effect of environmental conditions on photosynthesis \citep{monteith_solar_1972}. \ola{Usually} biome-specific relationships are established from empirical observations of GPP and APAR \citep[e.g.,][]{running_continuous_2004}, \ola{but \citet{Zhang2018} found that the expression of LUE based on PAR absorption by canopy chlorophyll tends to converge across biome types}.  

PBMs rely on more rigorous formulations than LUE models \citep[e.g.,][]{zhang_evaluating_2012}, but they have the disadvantage of complexity and uncertainty of their parametrization.
Although these input parameters for PBMs are interpreted as being more physical and biologically meaningful, many of them may be unavailable or highly uncertain. 

On the other hand, the fundamental assumptions underlying LUE models --that plant canopies behave like a big single-leaf, 
and their LUE is independent of the directional nature of solar radiation and vegetation structure-- have been widely questioned already by \citet{pury1997simple} and continue to be discussed with support of flux data measurements \citep{gu_advantages_2002, zhang_effects_2011, propastin_effects_2012}.
Furthermore, it is unclear how well these empirical relationships hold for spatial and temporal scales beyond those used to derive them, and how they might change under altering environmental conditions \citep[e.g.,][]{xin_multi-scale_2015}. 
The most widely used LUE model is applied in the Moderate Resolution Imaging Spectroradiometer (MODIS) GPP product, MOD17 \citep{running_continuous_2004}, {currently available (Collection 6)} globally at 8-day and {500~m resolution \citep{s._running_mod17a2h_2015}}. v
Despite general good performance of the model, evaluation of MOD17 for crops showed that it usually underestimates GPP for certain crops, for example soybeans and maize \citep[e.g.,][]{turner_site-level_2005,peng_remote_2012}.

This can be partly explained by neglecting
the high heterogeneity of different crop types 
and a coarse spatial resolution, which does not allow {separating} observations of individual fields, as well as different irrigation and fertilization practices that are important for crop performance \citep{zhang_evaluating_2012}. The VPM GPP V20 dataset --a more recent global GPP product that utilized MODIS datasets together with a reanalysis climate dataset and a land cover classification-- was based on an improved LUE theory that uses the energy absorbed by chlorophyll \citep{zhang2017global}, and it's overall accuracy was relatively high, though it also underestimated cropland GPP (by $\sim$15\%). 

A third {approach} to GPP estimation from remote sensing data is based on linking GPP fluxes at flux tower locations with observations of large spatial fields from satellites adopting advanced statistical and machine learning (ML) algorithms that use input variables from climate reanalysis and satellite data products \citep{xiao_estimation_2008,jung_towards_2009,jung_global_2011,Tramontana_2016}, such as the Max Planck Institute for Biogeochemistry (MPI-BCG) GPP product \citep{jung_global_2011}. 
Such methods are powerful in application, but being essentially a statistical approach, they share with more simple empirical LUE models the disadvantage of lacking the capacity to extrapolate to different conditions \citep{beer_terrestrial_2010}.
{In addition, the dataset needed to train such ML approaches should be sufficiently representative and cover a wide range of conditions, which is difficult in general and especially at the start of new satellite missions when the collected data is limited.}

GPP was also previously estimated at a finer spatial resolution;   
 e.g. \citet{gitelson_remote_2012} assessed crop GPP with the Landsat data (spatial resolution of 30 m).
They used the concept of total crop chlorophyll content, based on evaluation of performances of twelve vegetation indices (VIs) for estimating GPP using ground-based measurements \citep{peng_remote_2012}.
 
However, as these approaches use only simple VIs, the increased number of bands in the Sentinel-2 satellites and ongoing advancements in vegetation and GPP models motivate a more sophisticated application of available reflectance bands and the development of more flexible and powerful GPP algorithms. 
In this work, we propose a hybrid approach for GPP estimation~\citep{Camps-Valls:2011:RSI,verrelst_spectral_2016} based on the combination of process-based radiative transfer models (RTMs) with Sentinel-2 spectral reflectance data through ML algorithms. 
Rather than retrieving the biophysical parameters accounting for the impact of canopy structure and leaf pigments on the harvest of light, we convert spectral reflectance and meteorological information into GPP directly using statistical ML methods, such as 
random forests and neural networks. It is important to emphasize that the training is performed on the modeled data, rather than flux tower GPP, {which allows us to simulate a broad range of conditions. This, as well as the use of all reflectance data (instead of derived products), makes the study different from purely data driven ML algorithms such as that of \citet{jung_global_2011} and LUE models like \citet{gitelson_remote_2012}. By adapting the same modeled data for different spectral characteristics of various instruments, our approach can remain consistent among multiple past and future satellites and still makes the use of all available bands. Furthermore, it also can be applied across the range of spatial dimensions, \ola{independently of the footprints of the reference data used for the training of empirical models.}}
For the RTM, we use the soil-canopy energy balance radiative transfer model SCOPE to simulate the reflectance spectra, the light distribution in the vegetation, and the GPP as a function of the vegetation structure. The SCOPE model  incorporates leaf model 
{Fluspect \citep{vilfan2016fluspect}}
and canopy RTM 4SAIL \citep{verhoef_unified_2007}, which {can also be used for retrieving vegetation variables, e.g.,} leaf area index (LAI) and chlorophyll-\textit{a} and \textit{b} content ($C_{ab}$), as well as fraction of APAR (fPAR) that is used as input to a number of LUE models \citep[e.g., ][]{S2ToolBox_atbd}. We focus our investigation on soybeans and other C3 crops. 
{However, the same approach can be used for C4 crops, after running simulations with appropriate biochemical settings (e.g., photochemical pathway, maximum carboxylation capacity, and temperature response).}

This paper is structured as follows: in Section~\ref{sec:materials} we first introduce the SCOPE model,
{the satellite and meteorological data, flux tower sites for which our approach was tested, as well as other GPP models. We also introduce the methods applied for the analysis of the simulated data and the ML methods used.}
In Section~\ref{sec:results}, we analyze relationships between vegetation parameters relevant for GPP modeling, 
\ola{as well as the relationships between the components of the various LUE models, based on broad SCOPE simulations.}
We also discuss the performance of machine learning algorithms for different vegetation parameters, and explain how and why we chose to model GPP. {Afterwards, we compare the results of our ML models applied to the satellite data with other GPP models and flux tower estimates. Finally, in Section~\ref{sec:summary}, we conclude our findings and give an outlook for future work.}

\section{Materials and methods}
\label{sec:materials}

\begin{figure}[t] 
   \begin{center} 
   \includegraphics[width=1.0\textwidth]{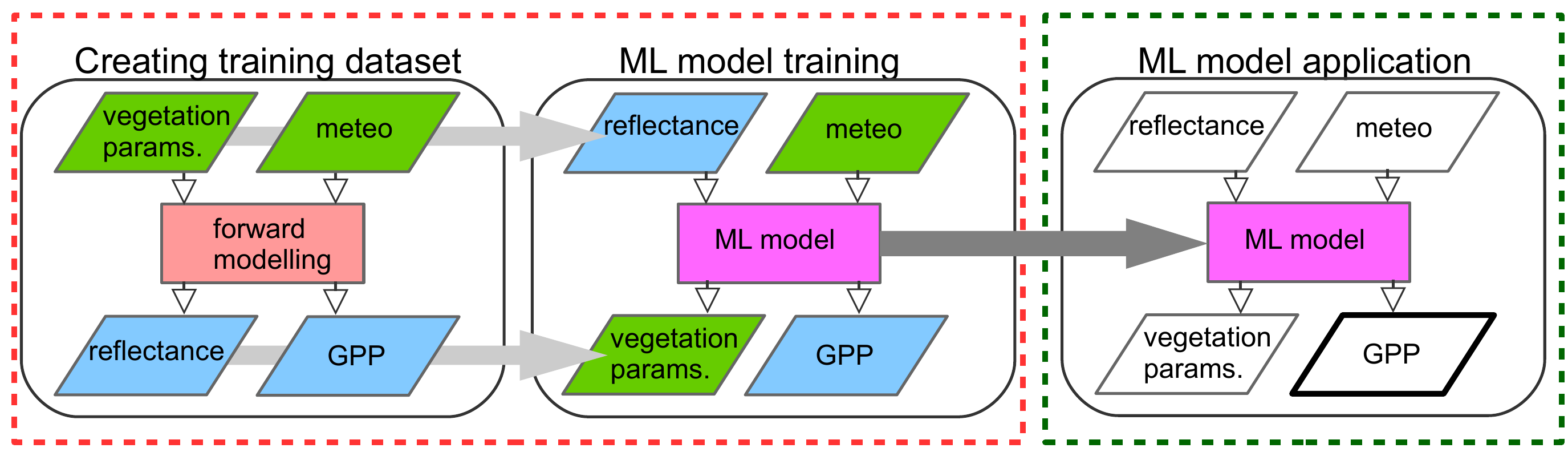}
     \caption{Flow chart of the processing chain applied in this work. The ML model is trained on the dataset created from SCOPE simulations, afterwards the ML model is applied to the {satellite and meteorological} data.} 
 \label{fig:flow_chart} 
   \end{center} 
\end{figure}

\begin{figure}[t] 
   \begin{center} 
   \includegraphics[width=1.0\textwidth]{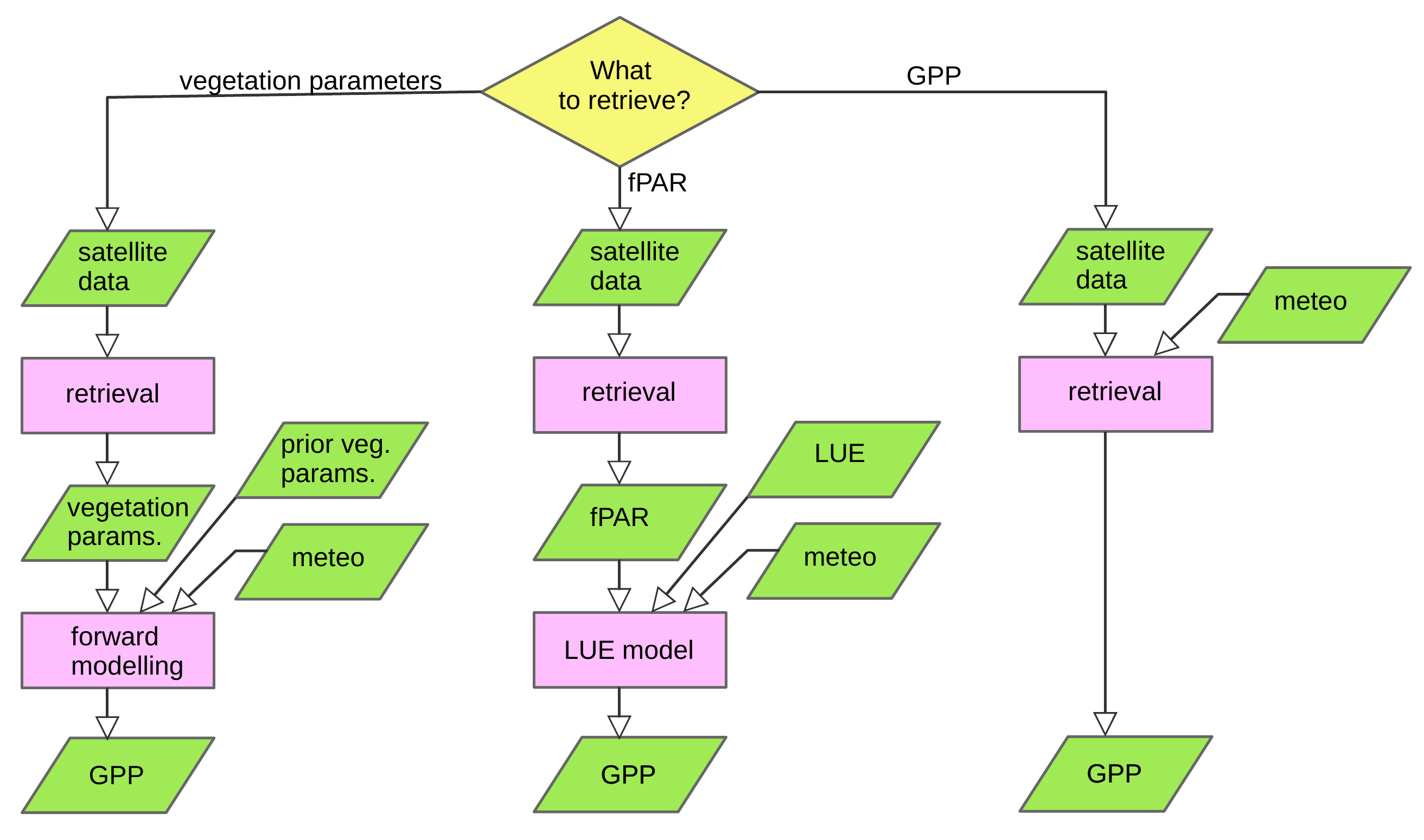}
     \caption{Workflows of three considered approaches to estimate GPP using data from SCOPE.} 
 \label{fig:possible_retrievals} 
   \end{center} 
\end{figure}

The overall process of creating and applying ML models for GPP estimation is schematically shown in Figure~\ref{fig:flow_chart}. 
{For creating the synthetic dataset we use the SCOPE model (Section~\ref{sec:scope}). Afterwards, the model is applied to the reflectance data of Sentinel-2 and Landsat 8 data (Section~\ref{sec:sen2} and Section~\ref{sec:landsat8}, respectively), and meteorological dataset GLDAS 2.1 (Section~\ref{sec:meteo}). 
Initially, we considered three different workflows to estimate GPP (Figure~\ref{fig:possible_retrievals}):
\begin{itemize}
	\item retrieving vegetation parameters from satellite data, then running the SCOPE model in a forward mode. In this case some vegetation parameters are estimated, while others have to be set a-priori;
	\item retrieving fPAR from satellite data, then applying LUE model;
	\item estimate GPP directly from satellite and meteorological data.
\end{itemize}
To analyze these approaches and finally decide on the most suitable method, we analyzed the synthetic dataset created with SCOPE. We performed global sensitivity analyses (Section~\ref{sec:gsa}), examined the relationships between fPAR, LUE and GPP, as well as tested ML algorithms for retrieving various parameters (Section~\ref{sec:ml}) using the modeled dataset. Eventually, we applied the ML model of GPP directly to satellite and meteorological data. 
We use data from flux tower sites (Section~\ref{sec:fluxsites}) for a feasibility test, and simple GPP models based on vegetation indices (Section~\ref{sec:gpp_vis}) for comparison with our ML model.
}

\subsection{The SCOPE model}
\label{sec:scope}

The SCOPE model is a vertical (1-D) integrated radiative transfer and energy balance model \citep{van_der_tol_integrated_2009}.
SCOPE calculates radiance spectra in the visible to thermal infrared range (\SI{0.4} to \SI{50}{\micro\metre}) as observed above the canopy, as well as the fluxes of water, heat and carbon dioxide.
SCOPE is continually updated, and recent significant improvements were introduced to link chlorophyll-\textit{a} vegetation fluorescence to the photosynthesis processes within the framework of the \textit{Photosynthesis Study} for the ESA FLEX mission \citep{mohammed_2014}.

SCOPE integrates radiative transfer and energy balance calculations at the level of individual leaves, as well as at the canopy level. 
The spectral transmittance and reflectance of the leaves are calculated with the Fluspect model \citep{vilfan2016fluspect}. 
Radiative transfer within the canopy is based on the 4SAIL model \citep{verhoef_unified_2007}. 
The leaf biochemical {processes} are based on \citet{collatz_physiological_1991} and \citet{collatz_coupled_1992} for C3 and C4 plants, respectively.
The geometry of the vegetation is treated in a stochastic way, where 
a probability of a leaf viewing in solar direction depends on the canopy parameters, 
and subsequently the different biophysical processes for sunlit and shaded components are considered.
To simulate photosynthesis, SCOPE requires inputs of meteorological forcing, vegetation structure parameters, leaf biophysical parameters, and optical and plant physiological parameters. {\ola{In the comparison of the simulated GPP (using Landsat data and locally measured weather data) to flux tower measurements,} \citet{Bayat2018} found a typical root-mean-square error of \SI{1.7}{\micro mol.s^{-1}.m^{-2}} (for GPP of about \SI{8}{\micro mol.s^{-1}.m^{-2}}, so about 20\% error), with an $r^2$ of 0.65 during a drought episode. Their relatively low $r^2$ was mainly due to the overestimation of GPP during the drought. They also showed that the accuracy can be improved by including thermal information.}
{More details on the SCOPE model can be found in \citet{van_der_tol_integrated_2009}.}

\begin{table}[]
\linespread{1.0}
 \scriptsize
 \centering
\caption{List of varied input parameters used in SCOPE model simulations. In this study we assumed uniform distribution of the input variables.}
\label{table:scope_params}
\resizebox{\textwidth}{!}{%
\begin{tabular}{*{6}{l}}
  \toprule
  Symbol & Parameter & Unit & Min & Max  \\
  \midrule
\textit{Leaf optical} & & & & \\
$C_{ab}$  & Chlorophyll-\textit{a} and \textit{b} content & \SI{}{\micro g.cm^{-2}} & 11 & 90 \\
$C_{ca}$  & Carotenoid content &  \SI{}{\micro g.cm^{-2}}  & 0 & 40 \\
$C_{ant}$  & Anthocyanins content &  \SI{}{\micro g.cm^{-2}}  & 0 & 40 \\
$C_{dm}$  & Dry matter content &  \SI{}{g.cm^{-2}}  & 0.0  & 0.05 \\
$C_{w}$  & leaf water equivalent layer &  \SI{}{cm} & 0.0 & 0.1 \\
$C_{s}$  & senescent material fraction & fraction & 0 & 0.9 \\
$N$  & leaf thickness parameter & - & 1 & 2.5 \\
\\
\textit{Canopy}  & & & & \\
LAI & Leaf area index & \SI{}{m^{2}.m^{-2}} & 0 & 9 \\
$h_c$ & vegetation height & \SI{}{m} & 0.1 & 2 \\
LIDF$_a$ & leaf inclination & - & -1 & 1\\
LIDF$_b$ & variation in leaf inclination & - & -1 & 1 \\
\\
\textit{Soil} & & & & \\
SMC & volumetric soil moisture content in the root zone & - & 0.01 & 0.7 \\
BSMBrightness & BSM model parameter for soil brightness & - & 0.01 & 0.9 \\
BSMlat & BSM model parameter 'lat' & - & 20 & 40 \\
BSMlon & BSM model parameter  'long'& - & 45 & 65\\
\midrule
\textit{Geometry} & & & & \\
SZA & solar zenith angle & degree & 0 & 85\\
\\
\textit{Meteorology} & & & & \\
$R_{in}$ & broadband incoming shortwave radiation (0.4-2.5 um) & \SI{}{W.m^{-2}} & 0 & 1400 \\
$R_{li}$ & broadband incoming longwave radiation (2.5-50 um) & \SI{}{W.m^{-2}} & 0 & 400 \\
$T_a$ & air temperature & \degree C & -10 & 50\\
$p$ & air pressure & \SI{}{hPa}& 500 & 1030\\
$e_a$ & atmospheric vapour pressure &  \SI{}{hPa} & 0 & 125 \\
$u$ & wind speed & \SI{}{m.s^{-1}} & 0 & 25\\
  \bottomrule
\end{tabular}%
}
\end{table}

Here, we used the most recent SCOPE model release (version 1.70), in which optical coefficients used by the leaf model are consistent with the latest PROSPECT-D model \citep{feret_prospect-d:_2017}. 
In addition, a new soil spectral reflectance Brightness-Shape-Moisture model (BSM) \citep{verhoef2018hyperspectral}, which is based on the Global Soil Vectors of {\citet{Chongya2012}}, has been added as an alternative to providing an input soil spectrum.
{In the BSM model, dry soil spectra are approximated using the soil brightness (B), and ``lat" and ``long" parameters that define spectral shape effects, while SMC parameter accounts for the soil moisture impact on the dry soil reflectance spectrum (see Table~\ref{table:scope_params}).}
In addition, a biochemical routine has been updated so that the internal CO$_2$ concentration in the leaf is calculated iteratively.
 
We adapted the model to work in parallel computing within the Matlab environment, and customized the input and output of the model as follows:
 \begin{enumerate}
  \item Added GPP to the output data, since the default output of the model covers only net canopy photosynthesis (GPP minus leaf dark respiration). 
   \item Added an option to calculate leaf maximum carboxylation capacity ($V_{cmax}$) at 25{\degree}C ($V_{cmax}^{25}$), as a function of chlorophyll concentration $C_{ab}$, following \citet{houborg_satellite_2013}:
  \begin{equation}
V_{cmax}^{25} = 2.5294 C_{ab} - 27.34,
\end{equation} 
{where $V_{cmax}^{25}$ is in [\SI{}{\micro mol.m^{-2}.s^{-1}}]
and $C_{ab}$ in [\SI{}{\micro g.cm^{-2}}]}.
 \end{enumerate}

 \subsection{Sentinel-2 data}
\label{sec:sen2}

  \begin{figure}[t] 
   \begin{center} 
   \includegraphics[width=1.0\textwidth]{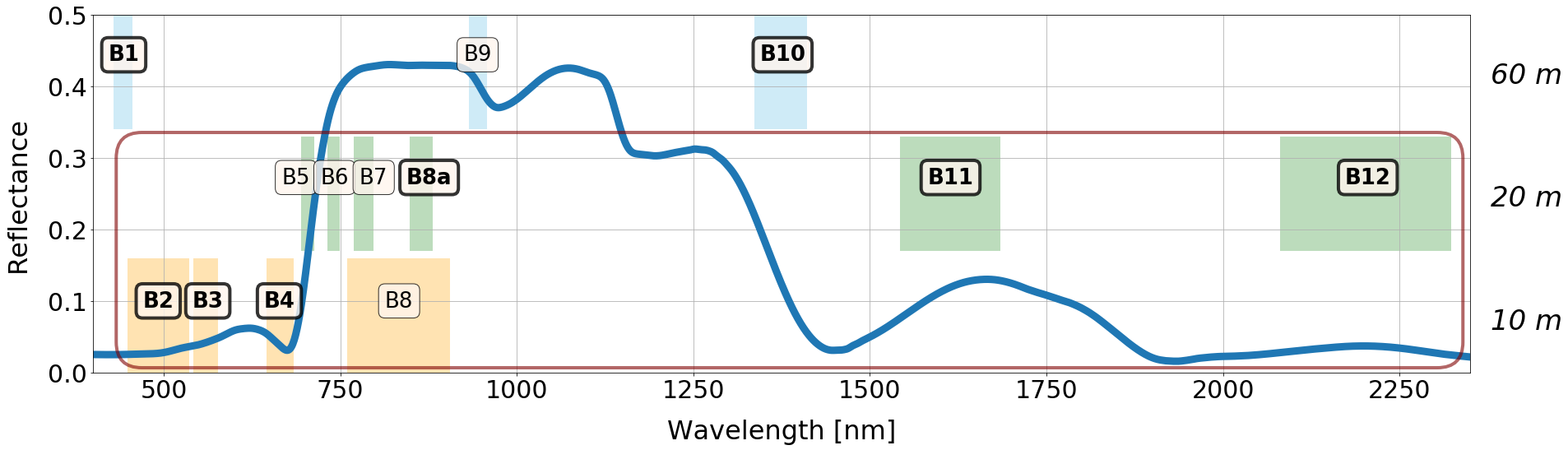}
\caption{\ola{Band settings of Sentinel-2A with respect to a typical vegetation reflectance spectrum. The bands in bold are those that overlap with Landsat 8 bands. The red rectangle encloses the bands used in this study.}}
 \label{fig:sen2} 
   \end{center} 
\end{figure} 
 
Sentinel-2 is a wide-swath, high-resolution, and multi-spectral imaging mission, supporting Copernicus Land Monitoring, including the monitoring of vegetation, soil covers and water bodies, as well as observation of inland waterways and coastal areas. The Sentinel-2 Multispectral Instrument (MSI) samples 13 spectral bands
spanning from the visible and the near infrared to the shortwave infrared (Figure~\ref{fig:sen2}), including two new spectral bands in the so-called red edge region (at 705~nm and 740~nm), which are very
important for retrieval of chlorophyll content \citep{clevers_remote_2013}. 
The spatial resolution varies from 10 m to 60 m depending on the spectral band with a 290 km field of view \citep{drusch2012sentinel}. 
Three bands at 60~m are mainly dedicated for atmospheric correction and cloud screening, which leaves ten bands aimed at \textit{land} surface observations. Currently, there are two Sentinel-2 satellites operating in tandem: Sentinel-2A was launched in June 2015, and Sentinel-2B launched in March 2017, which enables a revisit time of {less} than 5 days. 

We visually chose relatively cloud-free images over fields of interest (see Section~\ref{sec:fluxsites}) {for the years 2016-2017}.  
These images were atmospherically corrected using the Sen2Cor (version 2.4) algorithm, converting top-of-atmosphere (TOA) reflectance into top-of-canopy (TOC) reflectance \citep{dlr107381}. In addition, Sen2Cor delivered information on pixel {quality} (cloud, cloud shadow, etc.). We used this information subsequently to evaluate more precisely if the images were cloudy. In few cases, when available, we used atmospherically corrected TOC data directly. Obtained TOC bands (B2-B8, B8a, B11, B12) were re-sampled to a common resolution of 20~m using the SNAP toolbox. {We do not consider the effects of the resampling procedure, as we eventually calculate a mean value of the GPP over the whole fields.} 

\subsection{Landsat 8 data}
\label{sec:landsat8}

{Landsat 8, a NASA and USGS collaboration, is the latest of the Landsat series and 
was launched in February of 2013. 
Operational Land Imager (OLI), an instrument onboard the Landsat 8 satellite, 
has overall similar spectral coverage to Sentinel-2 (sharing six common bands, see Figure~\ref{fig:sen2}), but unfortunately does not cover as densely the vegetation red edge bands. The images of the Earth are collected with a 16-day repeat cycle, with a resolution of 30 m for bands of our interest \citep{Storey2016}.
}

{We used in this study atmospherically corrected surface reflectance from Landsat 8/OLI (USGS Landsat 8 Surface Reflectance Tier 1) from the Google Earth Engine (GEE) platform \citep{gorelick_google_2017}. These data have been atmospherically corrected using LaSRC \citep{Vermote2016} and include a cloud, shadow, water and snow mask produced using CFMASK \citep{Foga2017}, as well as a per-pixel saturation mask. We only used pixels which were marked \textit{clear} by pixel quality attributes generated from the CFMASK algorithm.
}

\subsection{Meteorological data}
\label{sec:meteo}
We used the meteorological data from Global Land Data Assimilation System (GLDAS) 2.1 that ingests satellite and ground-based observational data products to generate optimal fields of land surface states and fluxes \citep{rodell_global_2004}.  
GLDAS has been developed jointly by National Aeronautics and Space Administration (NASA) Goddard Space Flight Center (GSFC), and the National Oceanic and Atmospheric Administration (NOAA) National Centers for Environmental Prediction (NCEP).  
It extends from the year 2000 to present with about one month latency and is updated monthly. 
Choosing this dataset as meteorological input was also motivated by its availability {on} GEE, which we plan to use in future for applying our GPP model globally. 
We used 3-hourly GLDAS-2.1 land surface model data available through GEE with a resolution of 0.25\degree. 
{The data was exported for the dates of interest (availability of Sentinel-2 data) for the years 2016-2017. For Landsat 8 application, the GLDAS-2.1 data was used directly in GEE to estimate GPP. We did not perform any spatial interpolation, and used the meteorological data directly from the grid cells covering the chosen fields.}

\subsection{Flux tower sites}
\label{sec:fluxsites}

We used data from four flux tower sites located in the USA (US-Ro 1, US-Ro 2, US-Ro 5, US-Ro 6) and one site in Germany, DE-RuS (SE\_EC\_001 in the TERENO data portal \url{http://teodoor.icg.kfa-juelich.de}
), {for the feasibility test of our GPP models.}
Information on the location of the sites and crop types grown there can be found in Table~\ref{table:flux_sites}.
The sites were chosen
based on the type of crop (soybeans and other C3 crops). The data was acquired for the dates of available Sentinel-2 observations for the years 2016 and 2017.
We obtained GPP data directly for dates of from the Principle Investigators of the sites, and integrated half-hourly data to daily GPP values, which were then used as the reference value for the validation of our GPP model.

\begin{table}[]
 \tiny
\centering
\caption{Details about the flux tower sites used in this study \citep{griffis_measuring_2004,Ney2017}.}
\label{table:flux_sites}
\resizebox{\textwidth}{!}{%
\begin{tabular}{*{5}{l}}
  \toprule
   Site ID & Lon (\degree W)& Lat (\degree N)& Period & Crops  \\
  \midrule
US-Ro1  & -93.0898 &   44.7143  & 2016  & soybeans \\ 
US-Ro2  & -93.0888 &  44.7288   & 2016  & Kura clover only \\ 
US-Ro5  &   -93.0576 & 44.6910  & 2017  & soybean  \\ 
US-Ro6  &  -93.0578 &  44.6946  & 2017  & wheat/Kura Clover \\ 
DE-RuS  &  6.4472 &  50.8659  & 2016 \& 2017  & \begin{tabular}{@{}l@{}}winter barley in spring 2016, a catch crop \\ mixture in fall 2016 and sugarbeet in 2017 \end{tabular}  \\ 

  \bottomrule
  
\end{tabular}%
}

\end{table}

\subsection{GPP estimated with vegetation indices}
\label{sec:gpp_vis}
{LUE models making use of} VIs and 
incident photosynthetically active radiation (PAR$_{in}$) to estimate GPP for crops, were also applied for a comparison with our model. 
In previous studies different VIs were tested using ground-based in situ reflectance measurements \citep{peng_remote_2012}, as well as Landsat \citep{gitelson_remote_2012} and MODIS data \citep{peng_remote_2013}. 
These studies suggested different equations for GPP models using various VIs as input. The ones that showed the best performance in these  studies were also tested here (cf. Table~\ref{table:Vis}). 
{We applied these approaches because they can be relatively easily adapted for our case (Sentinel-2 data, daily values), as compared to studies using MODIS data at 8-day temporal resolution and minimum 500 m spatial resolution \citep[e.g.][]{Wagle2015,Yuan2015,Zhang2014}.}
For VIs using a red edge band, we tested both red edge Sentinel-2 bands (B5 and B6), and eventually we chose B5, which led to higher correlation with flux tower GPP than using B6.
We calculated 45\% of daily integrated $R_\text{in}$ values {(following \citet{running2015daily})} from GLDAS 2.1 to obtain PAR$_{in}$ (even though the original equations were sometimes developed for PAR$_{in}$ and sometimes for potential PAR$_{in}$). 
In addition, since none of these equations was actually designed for the band setting of Sentinel-2, we also established a linear function of red edge NDVI (reNDVI) and PAR$_{in}$ using {the flux tower validation dataset and calibrating the function directly on this data (cf. Table~\ref{table:Vis})}.

\begin{table}[]
\centering
\caption{Summary of vegetation indices used in this study. $\rho_{\text{green}}$, $\rho_\text{red}$, $\rho_{\text{red edge}}$ and $\rho_\text{NIR}$ are reflectance in spectral bands of green, red, red edge and near-infrared spectral regions and the refer to Sentinel-2 bands B3 (560 nm), B4 (665 nm), B5 (705 nm) and B8 (842 nm), respectively.}
\label{table:Vis}
\resizebox{\textwidth}{!}{%
\begin{tabular}{*{6}{l}}
  \toprule
  Vegetation index (VI)  &  \begin{tabular}{@{}l@{}}VI\\ abbreviation\end{tabular}  & VI formula & 
   \begin{tabular}{@{}l@{}}GPP\\ ($x = VI\times PAR_{in}$)\end{tabular}
  & Reference \\
  \midrule
Red edge chlorophyll index & CI$_\text{red edge}$ &  $\rho_\text{NIR} / \rho_{\text{red edge}} - 1 $& $4.80ln(x) - 37.93$ & \citet{peng_remote_2012} \\
Green chlorophyll index & CI$_\text{red edge}$ &  $\rho_\text{NIR} / \rho_{\text{green}} - 1 $ & $5.13ln(x) - 46.92$  & \citet{peng_remote_2012} \\
 \begin{tabular}{@{}l@{}}Normalized difference  \\ vegetation index\end{tabular}
& NDVI  &  $(\rho_\text{NIR} - \rho_\text{red}) / (\rho_\text{NIR} + \rho_\text{red})$ & $2.07 x - 6.19$ & \citet{gitelson_remote_2012} \\
 \begin{tabular}{@{}l@{}}Green normalized   \\ difference vegetation index\end{tabular}
& greenNDVI  &   $(\rho_\text{NIR} - \rho_\text{green}) / (\rho_\text{green} + \rho_\text{green})$& $2.86x - 11.9$ & \citet{gitelson_remote_2012} \\
Enhanced vegetation index  & EVI & $2.5 (\rho_\text{NIR} - \rho_\text{red}) / (\rho_\text{NIR} + 6\rho_\text{red} - 7.5\rho_\text{blue} +1)$ & $2.26x - 3.73  $ &  \citet{peng_remote_2013} \\
\begin{tabular}{@{}l@{}}Red edge normalized   \\ difference vegetation index\end{tabular}
& reNDVI &   $(\rho_\text{NIR} - \rho_\text{red edge}) / (\rho_\text{NIR} + \rho_\text{red edge})$ & {1.61x - 1.75} & {this study} \\ 
  \bottomrule
\end{tabular}%
}
\end{table}

\subsection{Global sensitivity analysis}
\label{sec:gsa} 
Global sensitivity analysis (GSA) refers to a set of mathematical techniques aimed to analyze how the variation in the output of a numerical model can be attributed to variations of its inputs. Among others, GSA can be applied to evaluate the relative importance of each input variable in a model and can be used to identify the most influential variables affecting model outputs \citep{pianosi_matlab_2015}.

{Here, we used the PAWN method \ola{\citep[][the name derived from these authors names]{pianosi_simple_2015}}, which employs the entire model output distribution (cumulative distribution function, CDF) to quantify the sensitivity of the parameters and therefore it is applicable independently of the shape of the distribution. This is} in contrast to variance-based sensitivity analysis (VBSA) that uses only the output variance, which might be not sufficient if the output distribution is multi-modal or highly skewed (see \citet{pianosi_matlab_2015} for more details). In addition, PAWN can be tailored to focus on {particular ranges of the output}, {for instance extreme values \citep{pianosi_matlab_2015}}. 
In the PAWN method, the sensitivity of the model output to the parameters due to direct and interaction effects is estimated with a PAWN total sensitivity index ($T_{i}$).
The PAWN index has a range of variation between 0 and 1, {with larger values reflecting higher importance. An input can be concluded to be non-influential, when $T_{i}$ is below a threshold that depends on the chosen confidence level and the size of sample.} 
The parameter space was sampled using Latin Hypercube Sampling (LHS) \citep{mckay1979comparison}. In total, $N_u + N_c M_\text{PAWN} n_\text{PAWN}$ model runs are needed to approximate the total sensitivity index of all $M_\text{PAWN}$ parameters, where $N_u$ and $N_c$ are the sample sizes of unconditional and conditional CDFs, respectively, and $n_\text{PAWN}$ is the number of conditioning values of the model input. 
GSA was performed using the SAFE Toolbox \citep{pianosi_matlab_2015}.

\subsection{Machine learning models}
\label{sec:ml} 
ML techniques map the relationship between the input (e.g., reflectances) and output (e.g., GPP) by fitting a flexible model directly {to} the data.  Unlike parametric models that define an input-output mapping function, whose definition depends on a fixed set of parameters, the function in machine learning is typically non-parametric, nonlinear and very flexible. The weights of the model are fitted by using a training dataset (here provided by the forward modeling using SCOPE) in such a way that the model should perform well (i.e. provide accurate predictions) in a hold-out set, typically called validation or test dataset. 
\citet{verrelst_machine_2012} compared four ML regression algorithms as candidates for biophysical parameter retrieval for Sentinel-2 and -3 and showed that Gaussian Process (GP) regression gave the most promising results.
However, the main limitation of GP regression is the high computational cost for training and testing, as each test example has to be compared to all training samples \citep{quinonero-candela_unifying_2005}. 
Therefore, having in mind effective and global application of the developed models, we decided to eventually apply more efficient methods in terms of computational cost, such as neural networks (NNs) and random forests (RFs). 
 Neural networks learn a relationship between input and output variables by establishing a set of nonlinear units {(nodes with non-linear activation functions)} organized in layers and connected by weights {and biases} that are equivalent to the regression parameters of classical parametric models \citep{bishop1995neural}. They are a popular tool in the analysis of remotely sensed data \citep[e.g.,][]{mas_application_2008}, and have been already implemented in operational retrieval chains, including processing of Sentinel-2 data in the biophysical processor of the Sentinel Application Platform SNAP \citep{S2ToolBox_atbd}.
Random Forests (RFs) are ensemble methods, which means that a RF generates multiple estimators and aggregates their results. RFs can model complex interactions among input variables and are relatively robust with regard to outliers. They also have less parameters compared with NNs \citep{breiman_random_2001} and recently were successfully applied in remote sensing applications \citep{wang_estimation_2016,Tramontana_2016}.
We trained these ML algorithms for the retrieval of vegetation parameters and modeling GPP using the data originated from the SCOPE model only.

Two training setups were engineered for this purpose:
\begin{itemize}
\item Case 1: Retrieving canopy and leaf parameters \ola{as well as retrieving fPAR}. The input information is the reflectance data and the {solar zenith angle (SZA) of the satellite observation}.
\item Case 2: Directly retrieving GPP. Here the input information is the reflectance data, SZA of observation, meteorological conditions, and SZA of a given modeling time step {(which changes during the day, as opposite to the SZA of observation)}.
\end{itemize} 
Multiple models were trained to estimate GPP, using reflectance data at Sentinel-2 resolution with all ten spectral \textit{land} bands, but also a subset of bands that are common with Landsat 8 (i.e. B2, B3, B4, B8a, B11, B12, cf. Table~\ref{fig:sen2}).

The GPP models were eventually applied to satellite data and GLDAS 2.1 meteorological data \citep{rodell_global_2004} at {20 m spatial resolution (of the satellite data) and} 3~h temporal resolution (temporal resolution of the meteorological dataset), {for four temporal points per day (when incoming shortwave radiation was above zero). These values were then integrated to obtain daily values, which is a typical scale at which GPP from remote sensing data is evaluated.} 
To compare data with flux tower measurements, we calculated daily GPP for the fields by taking average of Sentinel-2 (or Landsat 8) pixels within each field. 

{The ML regressions were applied in Python and were built using the scikit-learn toolkit \citep{pedregosa_scikit-learn:_2011}. If not mentioned otherwise, the settings of the ML models were set to \textit{default} and the random state to 1.}
\subsection{Modeling set-up}
\label{sec:model_setup}
We run multiple sets of SCOPE simulations to perform GSA (Section~\ref{sec:gsa}), and train ML algorithms (Section~\ref{sec:ml}). 
Since the SCOPE model has a large number of input parameters, we tried to limit the number of considered parameters by building up on a recent study by \citet{verrelst_global_2015}, 
in which the driving parameters for reflectance and SIF were investigated.

First of all, we focused on a number of vegetation and soil parameters that could be potentially retrieved from Sentinel-2 data.
We varied 15 leaf, canopy and soil parameters (see Table~\ref{table:scope_params}) assuming a uniform distribution, while keeping other parameters constant.  
{We used a uniform distribution for overall simplicity and generalization, and to obtain even performances over the whole range of the input variation.  
Furthermore, the variables were considered independent. We acknowledge that other distributions and assumptions might have led to a different performance of our ML models. However, since we wanted to focus on the concept of our approach, further optimization of ML algorithms is beyond the scope of this study, especially when considering a small validation dataset.}

The value of $V_{cmax}^{25}$ (leaf maximum carboxylation capacity at 25{\degree}C) was set constant ($V_{cmax}^{25}$ = \SI{100}{\micro mol.m^{-2}.s^{-1}}), or varied as a function of $C_{ab}$. The constant value of \SI{100}{\micro mol.m^{-2}.s^{-1}} was chosen following \citet{zhang_estimation_2014}, who applied it for SCOPE simulations for soybeans, and is commonly estimated for C3 crops \citep{wullschleger_biochemical_1993,kattge_quantifying_2009}.
{Since we focused on the vegetation and soil parameters first, the meteorological} conditions were set to default SCOPE values {($R_{in}$=\SI{600}{W.m^{-2}}, $T_a$ = 20\degree C, $R_{li}$=\SI{300}{W.m^{-2}}, $p$=\SI{970}{hPa}, $e_a$=\SI{15}{hPa}, $u$=\SI{2}{m.s^{-1}}), as well as the SZA (30\degree)}.
{Regarding the geometry of observations, we used the constant values of \textit{the observation zenith angle} (0 \degree) and \textit{the azimuthal difference between solar and observation angle} (90 \degree), since both Sentinel-2 and Landsat 8 have a relatively narrow field of view as well as quasi-nadir observations. 
}
We point out that LIDF$_a$ and LIDF$_b$ parameters were not independently sampled, but {instead we used} their sum (LIDF$_{a+b}$) and their difference (LIDF$_{a-b}$). This is motivated by the fact that their (LIDF$_a$ and LIDF$_b$) values must be chosen such that the sum of their absolute values equals to (or is smaller than) one, and therefore these parameters are not independent \citep{verhoef1998theory}. 
As a solution we chose LIDF$_{a+b}$ and LIDF$_{a-b}$ to vary independently between -1 and 1, and based on their values we calculated LIDF$_a$ and LIDF$_b$.
For these settings, we run the PAWN analyses, where $N_u$, $N_c$ and
$n_\text{PAWN}$ were set to 1000, 400 and 30, respectively, which equaled to 181000 simulations. 

{In order to train the ML GPP model, we additionally varied in simulations meteorological parameters and solar zenith angle (SZA).} 
To cover a large range of vegetation and meteorological conditions, the variable ranges were based on previous studies that performed satellite retrievals and global {sensitivity} analysis of SCOPE data \citep{verrelst_global_2015, zhang_estimation_2014}. All 22 parameters (see Table~\ref{table:scope_params}) were varied assuming uniform distribution (we run in total 177000 simulations).  
{These simulations were then re-run with different SZAs.
This allows us to account for different SZAs at a given time step of GPP modeling (as opposed to the SZA during the Sentinel-2 observation).} The SZAs in this scenario were chosen randomly between 0\degree and 85\degree. In this final training dataset, reflectance, meteorological data and SZA of observation are based on the original dataset, while GPP and SZA at a given time step are based on the re-run simulations.

{For some combinations, the energy balance has not converged without adjusting the maximum iteration number or the maximum accepted error in the energy balance.  These cases were not included in the training dataset for the ML model.}
This led to underrepresentation of cases with small LAI.
To resolve this problem, we additionally performed a subset of calculations, where the value of LAI was set to 0.001, while all other variables varied as before (which led to a number of 10700 additional simulations).
For these cases, regardless whether simulations converged, we simply assigned the value of GPP to zero in all these scenarios. This subset was afterwards included in the training dataset to represent conditions of a very small (almost zero) LAI.

\section{Results and Discussion}
\label{sec:results}
 
This section gives empirical evidence of the performance of the proposed scheme for GPP estimation. We start the analysis by exploring the relative relevance of parameters using a sensitivity analysis approach. Then we examine the relationship between  the components of the LUE model. After this analysis we provide quantitative results of GPP estimations using machine learning methods for the training and validation datasets. Results are then further validated for some selected flux towers and crops.
 
\subsection{Predictor variables of GPP}
\label{sec:predictor_gpp}

  \begin{figure} 
   \begin{center} 
   \includegraphics[width=1.0\textwidth]{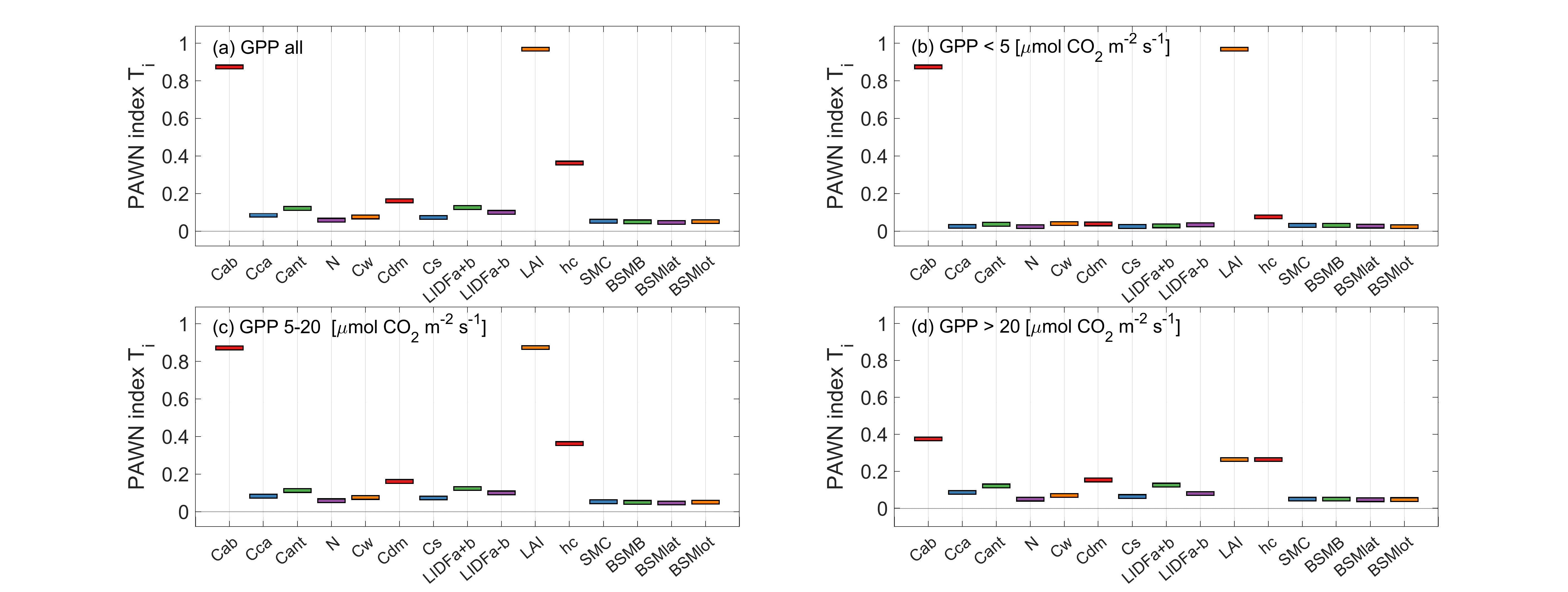}
     \caption{PAWN indices for GPP for simulations with $V_{cmax}^{25}$  dependent on $C_{ab}$ for (a) all data, or three GPP sub-ranges: (b) small ($<$\SI{5}{\micro mol.CO_2.m^{-2}.s^{-1}}); (c) medium (5-20 \SI{}{\micro mol.CO_2.m^{-2}.s^{-1}}); and (d) high ($>$\SI{20}{\micro mol.CO_2.m^{-2}.s^{-1}}). The boxes represent single values estimated for each input parameter.} 
 \label{fig:pawn_gpp_pawn2} 
   \end{center} 
\end{figure} 
 
We applied GSA to identify the most influential variables affecting GPP. {We focused} on the vegetation and soil parameters that can be potentially retrieved from the satellite data. {Such an analysis should help to decide which vegetation parameters ought to be estimated in order to accurately estimate GPP with the SCOPE model. We used here a dataset, where $V_{cmax}^{25}$ was varied as a function of $C_{ab}$, while meteorological conditions were constant.} 
The results of PAWN indices computed over the whole output range are shown in Figure~\ref{fig:pawn_gpp_pawn2}a. 
LAI was the most influential input ($T_{\text{LAI}}$ of 0.97), followed by $C_{ab}$ ($T_{h_c}$ of 0.87), while other parameters were much less influential ($T_{i} < 0.5$).
In addition, we performed the PAWN analysis in the following three GPP sub-ranges: small ($<$\SI{5}{\micro mol.CO_2.m^{-2}.s^{-1}}), medium (5-20 \SI{}{\micro mol.CO_2.m^{-2}.s^{-1}}) and high ($>$\SI{20}{\micro mol.CO_2.m^{-2}.s^{-1}}). The influences of input parameters vary substantially for different sub-ranges.
While LAI and $C_{ab}$ turn out to be the most influential parameters predominantly for small and medium GPP, other variables (and especially $h_c$) are also influential for high GPP (Figure~\ref{fig:pawn_gpp_pawn2} b-d). 
LAI is the parameter that in general controls the presence and abundance of vegetation, and hence has a dominant role in determining GPP.
The high influence of $C_{ab}$ is due to it's role in capturing light used for synthesis, but also because $V_{cmax}^{25}$ is set as a function of $C_{ab}$; 
$h_c$ is used in the model to calculate the {roughness length for the momentum of the canopy displacement height, which in turn has an effect on the leaf temperatures} and the gradients of water and CO$_2$ between the leaf surface and the atmosphere. As a result, we observe a high influence of $h_c$ on GPP.
Eventually, for high GPP, most of vegetation parameters become relevant to a certain degree (Figure~\ref{fig:pawn_gpp_pawn2}d).
{For example, other leaf pigments (content of carotenoid ($C_{ca}$) and content of anthocyanins ($C_{ant}$)), dry matter and canopy geometry parameters have all a stronger influence for medium and high GPP than for low GPP.}
Optically active leaf components 
compete with each other for light to absorb, while canopy geometry alters the relationship between sunlit and shaded leaves. 
The least influential {vegetation} variables are the leaf thickness parameter ($N$), leaf water equivalent layer ($C_w$)  and senescent material fraction ($C_s$). 
Soil parameters have no influence in all cases. This is due to fact that the effect of  soil properties on photosynthesis is not parameterized in the model. 
The ranges of the input variables were chosen to be very broad in order to cover many various scenarios (see Table~\ref{table:scope_params}), but at any given time and place, their actual seasonal and daily variability can be very different. 
{However, since it is very difficult to determine with certainty what is exactly probable for a global scale, we opted for including rather too much variability than too little.}
The description of the variability of each input parameter, including their range and distribution, can significantly affect the GSA results. For example, considering only a small range of pigment concentrations, could decrease their influence on the output parameters as seen in the GSA. On the other hand, the parameter that has overall smaller influence can be sometimes the most important one if it changes more drastically than other parameters.
{
{The} vegetation parameters {(LAI and $C_{ab}$)} that emerged as the most influential for estimating GPP, are exactly the ones that are often retrieved from remote-sensing data. 
However, more leaf and canopy parameters are important for precise calculation of GPP, especially for high GPP (which is often the case for crops, on which we focus in this study). 
Therefore, in case of using the SCOPE model, multiple further input variables should be estimated for running the model in the forward mode.  
Many of the variables are more difficult to estimate, and must be therefore assumed a priori, often with high uncertainties.}
{However, if these assumption are not well constrained, it makes the global applications of such a complex model challenging.} 

\subsection{Relationships between APAR and GPP}
\label{sec:predictor_all}

  \begin{figure} 
   \begin{center} 
   \hspace*{-12mm}\includegraphics[width=1.2\textwidth]{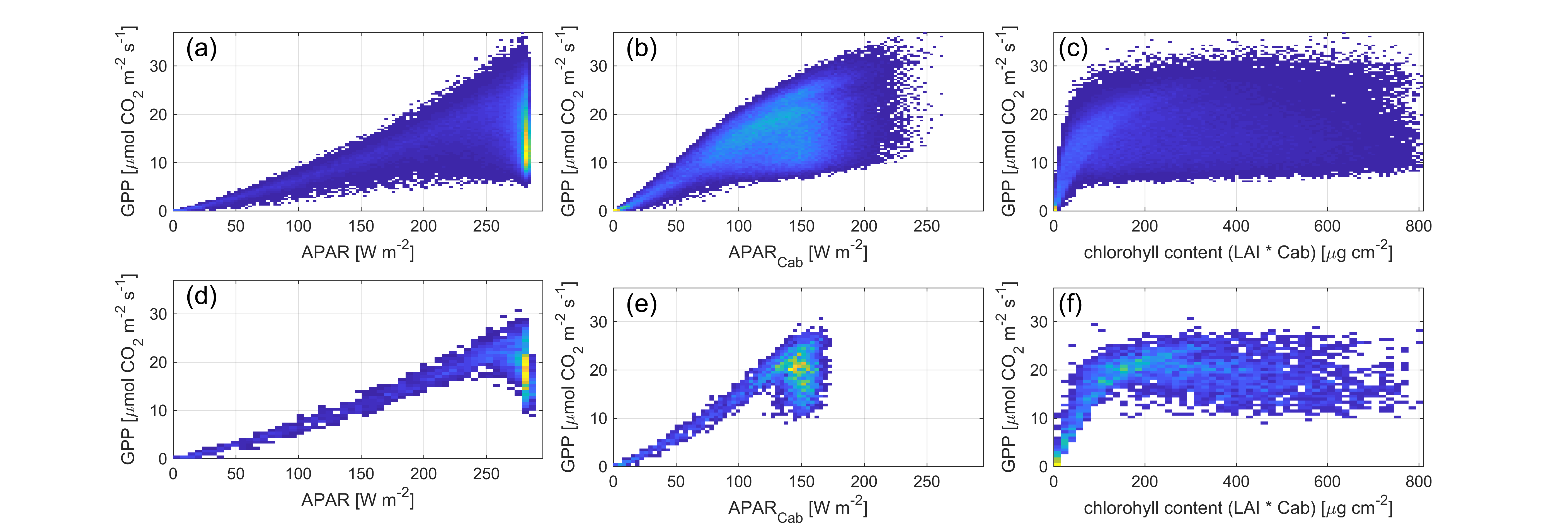}

     \caption{Relationships between GPP and (a,d) APAR, (b,e) APAR$_{\text{Cab}}$ and (c,f) {canopy} chlorophyll content (LAI$\cdot$$C_{ab}$) for (top panel) the whole dataset and (bottom panel) a selected subset ($C_{ca}$ 15-35\% of $C_{ab}$, $C_{ant}$ 30-60\% of $C_{ab}$, $h_c$ 1.2 - 1.6m). The data were calculated with SCOPE, assuming constant $V_{cmax}^{25}$ and constant meteorological conditions as default in SCOPE ($R_{in}$=\SI{600}{W.m^{-2}}, $T_a$ = 20\degree C, $R_{li}$=\SI{300}{W.m^{-2}},$p$=\SI{970}{hPa}, $e_a$=\SI{15}{hPa}, $u$=\SI{2}{m.s^{-1}}).} 
 
 \label{fig:hist_gpp_apar} 
   \end{center} 
\end{figure} 
 
\ola{We also considered using SCOPE to estimate GPP by means of applying a LUE model, in which the only retrieved parameter would be fPAR, while LUE would be assumed constant and adjusted only by meteorological conditions (Figure~\ref{fig:possible_retrievals}). In order to examine this application scheme, 
we analyzed the relationships between the components of the various LUE models (based on fPAR, fraction of PAR absorbed by chlorophyll, fPAR$_{Cab}$, or {canopy} chlorophyll content) as captured in SCOPE.} The relationships between APAR, APAR$_{Cab}$ (PAR absorbed by chlorophyll), {canopy} chlorophyll content {(LAI$\cdot$$C_{ab}$)} and GPP were examined for a simple case of constant $V_{cmax}^{25}$ and constant (default in SCOPE) meteorological conditions {($R_{in}$=\SI{600}{W.m^{-2}}, $T_a$ = 20\degree C, $R_{li}$=\SI{300}{W.m^{-2}}, $p$=\SI{970}{hPa}, $e_a$=\SI{15}{hPa}, $u$=\SI{2}{m.s^{-1}})}. { This was done in order to dismiss the impact of biochemical variability on LUE.}
\ola{Overall, obtained relationships are nonlinear, strongly scattered and heteroscedastic (Figure~\ref{fig:hist_gpp_apar}), which does not agree with the assumptions made in the foreseen LUE model, where LUE would only vary with meteorological conditions.}
These simulations were performed for a constant irradiance and, therefore, the light response is not as saturating as typical light response curves. Thus, all variability is due to changes in leaf and canopy properties.
However, it is important to keep in mind that this dataset includes strongly varied combinations. 
By limiting the data variability, for example by narrowing the relations between pigments and height of the canopy (e.g., $C_{ca}$ 15-35\% of $C_{ab}$, $C_{ant}$ 30-60\% of $C_{ab}$, $h_c$ 1.2 - 1.6 m, {chosen just as an example not to suggest any specific distributions}), these relationships become much more linear, with larger scatter of data points only for high APAR$_{Cab}$, APAR and {canopy} chlorophyll content. 
{
This agrees with the GSA analysis, which also showed that the impact of more parameters is larger for high GPP.
Since we use constant $R_{in}$ (\SI{600}{W.m^{-2}}), this high variability towards larger values can be due to the photosynthetic efficiency being mitigated in the excessive light by other parameters (e.g., LIDFa that controls the angular distribution and therefore the ratio of sunlit and shaded leaves). As compared to APAR, APAR$_{Cab}$ appears to be a better parameter for estimating GPP.} 
In the case of {canopy} chlorophyll content, the relationship saturates between 100 and \SI{200}{\micro g.cm^{-2}}.
\citet{gitelson_efficiency_2016} directly related GPP with {canopy} chlorophyll content and argued that GPP divided by incident PAR remained invariant, supporting the concept of an optimization of resource allocation \citep{goetz1999modelling,field1991ecological}.  
Since we used constant $R_{in}$ in our simulations, we can directly compare the shapes of our curves with \citet{gitelson_efficiency_2016}. 
It is remarkable that \citet{gitelson_efficiency_2016} also reported that GPP was very sensitive to {canopy} chlorophyll content up to \SI{150}{\micro g.cm^{-2}}, and not so much for {canopy} chlorophyll content above. Still, our results are more scattered and show a higher variability of GPP. However, since we used only a synthetic dataset, we can neither  support nor refute their functional convergence hypothesis.

\ola{In general, these results suggest that LUE can change due to differences in leaf and canopy properties and that, by limiting the variability of the input variables, the SCOPE modelling results can converge into a more constant LUE. However, the confining of the input parameter distribution is not straightforward when all possible cases have to be accounted for. Nevertheless, as \citet{Zhang2018} found that maximum LUE (based on APAR$_{Cab}$) tends to converge across space and time, this approach could be further improved when the specific distributions of input parameters and their co-dependencies are investigated in detail.}

\subsection{Training of vegetation parameters and GPP models}
 
{Since before we were considering different workflows of estimating GPP using SCOPE (Figure~\ref{fig:possible_retrievals}), we also tested the performance of various ML algorithms on the data modeled with SCOPE (dataset with $V_{cmax}^{25}$ varying as a function of $C_{ab}$ and changing meteorological conditions).} 
{In addition to GPP models, we compared performances of ML algorithms retrieving vegetation biophysical parameters using the SCOPE data.}  
{The GPP model was trained using satellite and meteorology data, while the other considered ML algorithms for LAI, {canopy} chlorophyll content and fPAR$_{Cab}$, used only satellite data.}
We used synthetic data to both train and validate ML algorithms.  . 
Simulated samples from SCOPE were hence divided into training (85\%) and validation (15\%) subsets. 
For LAI, {canopy} chlorophyll content and fPAR$_{Cab}$ retrievals, we used all ten (B2-B8, B8a, B11, B12) Sentinel-2 bands {(reflectance output of SCOPE convolved to Sentinel-2 bands using spectral response functions)} and the SZA of the observation. For models estimating GPP we used Sentinel-2 bands, both SZAs (of the observation and the modeling step), and the meteorological data.
 
  \begin{figure}[h!] 
   \begin{center} 
   \includegraphics[width=1.0\textwidth]{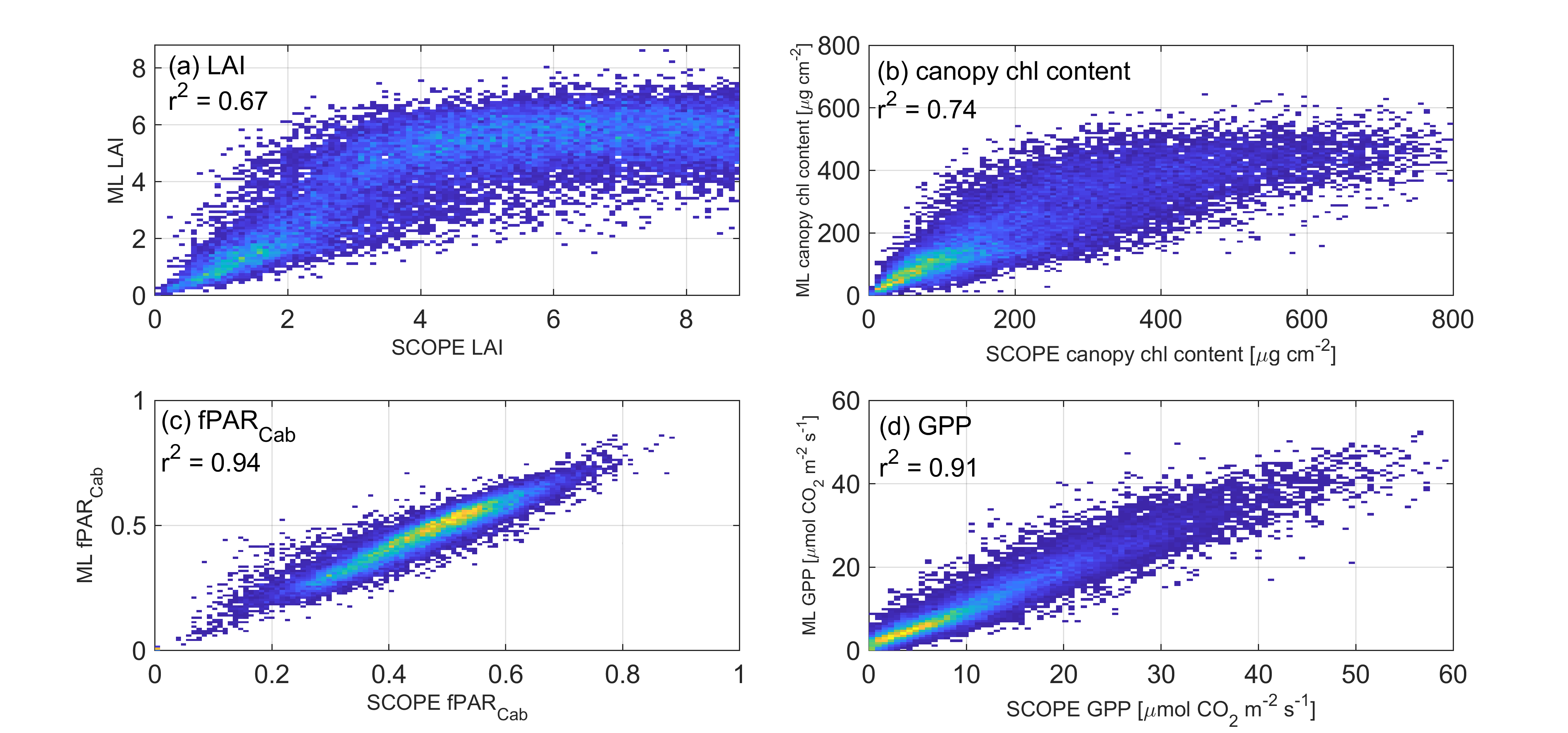}

     \caption{Performances of NN models for (a) LAI, (b) {canopy} chlorophyll content (LAI$\cdot$$C_{ab}$), (c) fPAR$_{Cab}$, and (d) GPP on the test dataset (the subset of SCOPE simulations).
}  
 \label{fig:laiccagpp_training} 
   \end{center} 
\end{figure} 

{We show results for NN, as RF gave similar results (data not shown).} 
First, to compare the performance of the NNs for different parameters, we built a NN for each output separately. 
{To minimize the effects of different atmospheric corrections and to at least partly harmonize the spectra across the sensors, for which we plan to apply our approach, we normalized the reflectance data to their spectral integral. In future, it would be optimal to use harmonized products, like e.g., \citet{Claverie2018}.}   
In addition, all inputs were normalized   
to fall in 
the range between zero and one (MinMax scaler).
For this comparison, we used a simple NN: two hidden layers with 12 neurons with rectified linear unit function (ReLU). We also tested briefly other network structures, but results were similar and are not shown.  

{The performance of NN models for validation subsets is shown in Figure~\ref{fig:laiccagpp_training}}
Our results show that retrieving vegetation parameters with a good accuracy is, in general, problematic.
Even LAI, which is a commonly retrieved parameter, is difficult to estimate with a certain accuracy. 
For example, LAI has similar (even though much stronger) effects on the reflectance to LIDFa. Overall, retrieving multiple parameters from a limited number of Sentinel-2 bands is by nature an ill-posed inverse problem, where a set of possible solutions could lead to a match between the measured and the simulated reflectance data. Therefore, additional prior information can be helpful to improve the solution \citep{combal_retrieval_2003}.

Our NN model of LAI showed a worse performance than the LAI retrieval that is implemented in SNAP, {which is also} using NN trained on the PROSAIL data \citep{S2ToolBox_atbd}.   
Their results were {validated} with an independent test dataset (simulated using the same radiative transfer model), which also showed a minor LAI underestimation (but only around 6) and a much better performance overall. 
In addition, a comparison of LAI derived from Sentinel-2 data using this SNAP algorithm with a small set of non-destructive (optical) field reference measurements showed a very good agreement, with $r^2$ of 0.83 \citep{vuolo2016data}. 
{
There are several issues that might lead to the poor performance of our NN model. First of all,  
we used a more recent version of the SCOPE model with a higher number of input parameters, which can make it more difficult to train NNs.} {Furthermore, not only a number of input parameters, but also their distribution differs among this study and \citet{vuolo2016data}, who used a Gaussian distribution. We used a uniform distribution and LHS in order to limit assumptions on underlying parameters as much as possible, but using a Gaussian distribution was shown to improve performance of the LAI retrieval in case of \citet{Verger2011}. However, since GPP retrieval performed satisfactorily and it is the main focus in this study, we did not test adding additional a priori assumptions. We also applied a normalization of the satellite spectra to their sum across all bands, which might lead to the loss of some information from the magnitude. Overall, this is an ill posed problem, and an algorithm specifically designed to estimate LAI may impose a number of conditions to regularize the problem, as compared to our approach where we focus on GPP estimation.}

{NN performed better for {canopy} chlorophyll content and fPAR$_{Cab}$ than for LAI (Figure~\ref{fig:laiccagpp_training}c and d), which agrees with the study of \citet{Verger2011}.}
For {canopy chlorophyll content}, we obtained much better performance ($r^2$ of 0.74, as compared to $r^2$ of 0.67 for LAI), with only minor underestimation above \SI{500}{\micro g.cm^{-2}}. For fPAR$_{Cab}$, NN produced much better results ($r^2$ of 0.94), which shows that even {though} here we were not able to retrieve leaf and canopy properties accurately, more general characteristics of absorbed radiation can be retrieved really well. {Therefore, at least the fPAR element of the LUE model can be well observed using the Sentinel-2 data.}
 
The GPP model performed much better than the LAI or {canopy} chlorophyll content retrievals, though similarly to fPAR$_{Cab}$, 
{with a very small bias (mean error = \SI{0.2}{\micro CO_{2}.s^{-1}.m^{-2}}) across the complete GPP range.  (Figure~\ref{fig:laiccagpp_training}).}  
This suggests that it may be possible and actually easier to directly estimate GPP than to first retrieve other vegetation parameters, which would be then afterwards used to estimate GPP by running the original model in the forward mode.  
Good performance of the GPP and fPAR$_{Cab}$ models also suggests that the important information is already available in the Sentinel-2 bands, despite the fact that an accurate retrieval of leaf and canopy variables is very challenging (e.g., Figure~\ref{fig:laiccagpp_training}). {However, using retrieved fPAR$_{Cab}$ in the LUE model is also not straightforward, as according to our previous analysis, LUE is not constant in SCOPE (Section~\ref{sec:predictor_all}).} 

Therefore, we chose to apply the ML model of GPP directly to the satellite data, instead of performing middle-step retrieval of vegetation parameters followed by re-running the original model in the forward mode.
Our method makes the best use of the complexity of the process-based model (here SCOPE) in conditions of limited information about the system that we usually have, as it combines deep understanding of photosynthesis as implemented in the original model with the ML algorithms that are appropriate for the application to remote sensing data on a global scale.
In addition, the algorithm design makes it easy to adjust or improve it, when the new version of the model is available - the data for training would have to be re-calculated and the algorithm re-trained, but it could be thereafter directly applied in an identical manner.
Furthermore, ML models can be trained on data from the same model but with different spectral settings, which allows a global application across a range of different satellites while still being based on the same model.

\begin{table}[]
 \scriptsize 
\centering
\caption{{Performance ($r^2$) of different ML models on training and test datasets (both with SCOPE), as well as on the validation (val.) data from the flux tower sites. For five different NN models we varied the number and size of hidden layers, as shown in the table.}}
\label{table:ML_results}
\resizebox{\textwidth}{!}{%
\begin{tabular}{c*{5}{C}}
  \toprule
  \smash{\raisebox{-.55\normalbaselineskip}{}} & 
        \multicolumn{5}{c}{NN (hidden layers)} \\
  \cmidrule(lr){2-6}	         
        	    & \begin{tabular}{@{}c@{}} \#1 \\ (12,12) \end{tabular}  & \begin{tabular}{@{}c@{}} \#2 \\ (20,20)  \end{tabular} & \begin{tabular}{@{}c@{}} \#3 \\ (20,12)  \end{tabular}  & \begin{tabular}{@{}c@{}} \#4 \\ (12,12,12) \end{tabular}  & \begin{tabular}{@{}c@{}} \#5 \\ (40,20,12) \end{tabular}   \\

  \midrule
GPP r$^2$ train &  0.92 & 0.93 & 0.92 & 0.93 & 0.96 \\
GPP r$^2$ test &   0.92 & 0.94 & 0.92 & 0.93 & 0.95 \\
LAI r$^2$ test &  0.58 & 0.62 & 0.62 & 0.59 & 0.68 \\
GPP r$^2$ val. &  0.86 & 0.88 & 0.92 & 0.89 & 0.91 \\
GPP RMSE val. &   1.72 & 1.66 & 1.38 & 1.51 & 1.41 \\
  \bottomrule
\end{tabular}%
}

\end{table}

{Eventually, we also added LAI as an output to the ML models trained primarily to retrieve GPP, since LAI is the most influential parameter for GPP according to GSA.
Although overall the performances of both architectures were similar, we obtained a small improvement in the performance of ML for GPP in case of our final model settings (of 0.01 for r$^2$ and RMSE=\SI{0.2}{gC.d^{-1}.m^{-2}} for the training and test synthetic datasets).}
We considered different structures of the final ML models,
{and tested their performances on the training, testing and validation datasets (Table~\ref{table:ML_results} and Table~\ref{table:ML_results_RF}).}
The results for both NN and RF were overall very similar. However, RF was in our case slightly affected by overfitting, since the determination coefficients were higher for the training than for the testing datasets.
This was not the case for NNs, {which had similar performance for training and testing datasets.} 
Therefore, we focus here on NN models, while results for RF are shown in {A3}.
{However, our approach in general does not rely on any specific ML method.}

For NNs, we varied the number of hidden layers, the number of neurons per layers, the batch size {(the number of training examples utilized in one iteration)} as well as activation functions for hidden layers. After examination, we decided to use a batch size of 32, and ReLU as the activation function because of its good performance and low computational cost. 
{Eventually, before finally applying ML algorithms to satellite data, we re-trained these ML models with the whole available SCOPE dataset.}
As for the numbers of hidden layers and neurons per layers, many different settings gave similarly good results for the synthetic dataset, so we ultimately compared the performances of {five chosen architectures with the flux tower data  (r$^2$ between  0.86 - 0.92, cf. Table~\ref{table:ML_results})}.
Eventually, we chose the NN model that performed best on the flux tower validation dataset. The best settings chosen for NN  
were afterwards applied to another ML model of GPP, {for which we used only Sentinel-2 bands shared with Landsat 8}.

 \begin{figure} 
   \begin{center} 
   \includegraphics[width=0.8\textwidth]{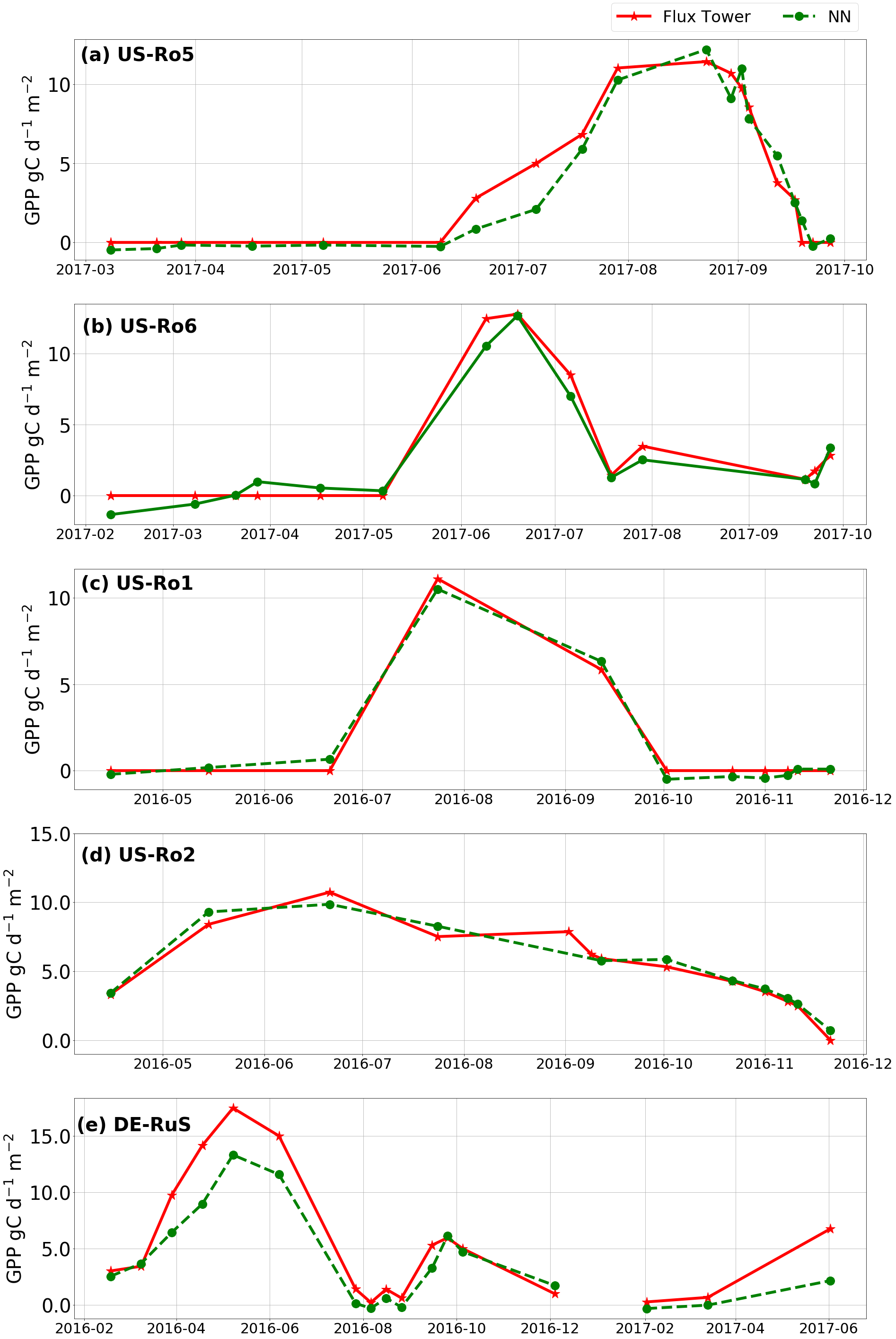}

     \caption{Time series of GPP estimated from flux towers (red), and modeled with NN (green). Data points correspond to the days for which a clear Sentinel-2 image was available.} 
 \label{fig:gpp_retrieval_1a} 
   \end{center} 
\end{figure}

 \begin{figure} 
   \begin{center} 
   \includegraphics[width=1.0\textwidth]{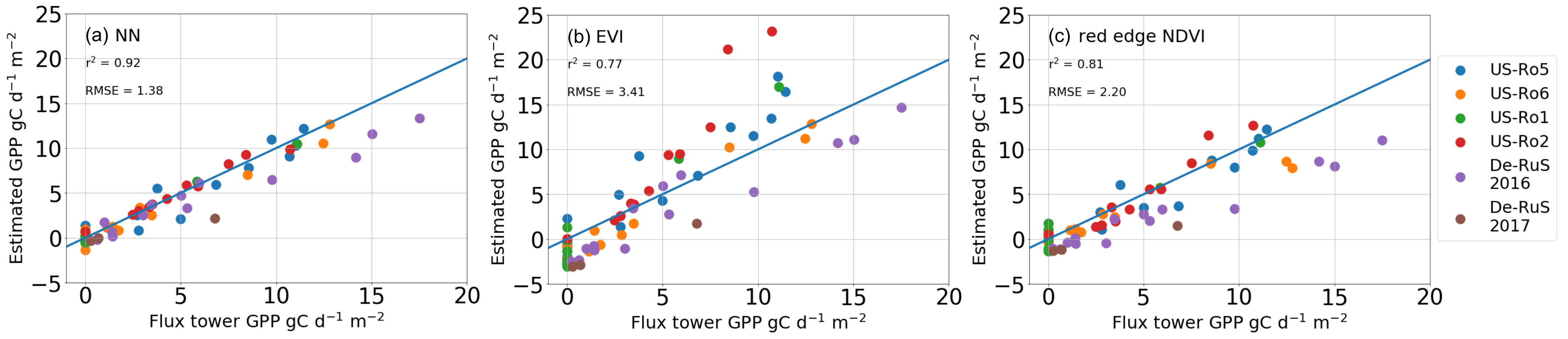}

     \caption{{Relationships between daily flux tower GPP and GPP estimated using (a) NN, (b) EVI and (c) reNDVI. Functions used for VIs are shown in Table~\ref{table:Vis}. The straight line shows a 1:1 relationship.}} 
 \label{fig:scatter_gpp} 
   \end{center} 
\end{figure}

\subsection{Application to Sentinel-2}
 \label{sec:retrievals_gpp_application}

The GPP model was applied to {the processed Sentinel-2 data}. 
The comparison of resulting time series of GPP measured at flux towers and estimated GPP are shown in Figure~\ref{fig:gpp_retrieval_1a}, and the overall results are compared in scatter plots in Figure~\ref{fig:scatter_gpp}. 

The NN model captures well the seasonal dynamics of the GPP, both in terms of the magnitude, as well as of the phenology. The good performance of the model is additionally confirmed by a strong linear relationship that was established for all seasons ($r^2 = 0.92$). The models successfully estimated GPP {also outside of the growing season}, and precisely tracked the emergence and senescence/harvest.

It must be stated that these results are for clear-sky data. Spectral reflectance data are affected by clouds, which is not accounted for in our statistical training and introduces errors in GPP models. In the first step of the selection of cloud-free days, Sentinel-2 cloud-free images were chosen visually. However, additional cloud check (based on the Sen2Cor classification of pixel into cloud, snow, shadow, etc.), showed that we included few days when the fields were covered by thin clouds, or partly by clouds and/or cloud shadows. These days show worse results {(underestimation for most cases, cf. Figure~\ref{fig:gpp_ro1_rf_nn}), which stresses the importance of atmospheric correction and scene classification.} 

Our ML algorithms performed better as compared to VI models (Figure~\ref{fig:scatter_gpp}). However, the reNDVI model, which was {fitted directly to} the flux tower dataset, also yielded good results ($r^2$=0.77 and RMSE=\SI{4.41}{gC.d^{-1}.m^{-2}} for EVI, and $r^2$=0.81 and RMSE=\SI{2.16}{gC.d^{-1}.m^{-2}} for reNDVI). 
{We note that for VI models, using the red edge band in the reNDVI indeed improves GPP estimation as compared to EVI, and such  
LUE models are already widely used and show an overall good performance \citep[e.g.,][]{gitelson_remote_2012,peng_remote_2013,Zhang2014,Wagle2015,Yuan2015}. However, we see the strength of our method not in directly outperforming empirical models, but in its potential to be used across a range of instruments with different spectral and spatial characteristics, and for a range of different conditions that might not be captured by empirical models.} A good performance of our model was obtained despite using no empirical information.
First of all, it proves that the SCOPE model performs well and that it is a good and a reliable tool for coupled radiative transfer and biochemical modeling and therefore for relating reflectance data with GPP, also without having local information from the site. 
Accordingly, a possible future global application looks very promising.

 \begin{figure} 
   \begin{center} 
   \includegraphics[width=1.0\textwidth]{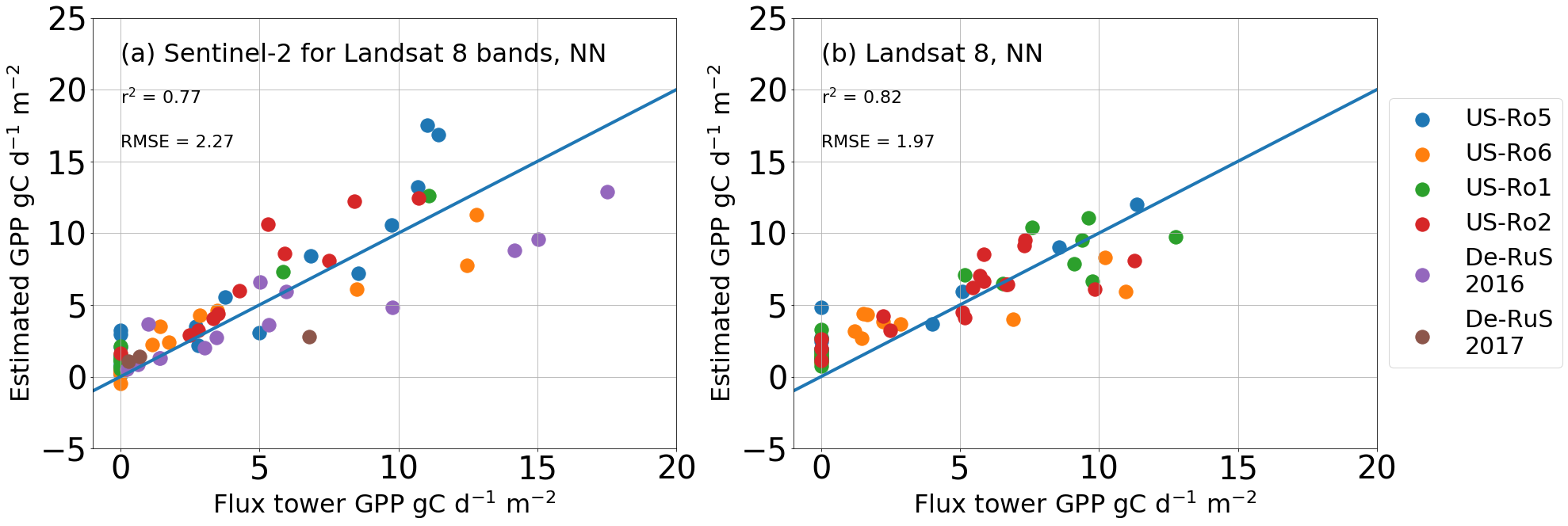} 

     \caption{Relationship between daily flux tower GPP and GPP estimated using (a) the subset of Sentinel-2 spectral bands that are also available in Landsat 8, (b) Landsat 8 data. The straight line shows a 1:1 relationship.} 
 \label{fig:scatter_gpp_landsat} 
   \end{center} 
\end{figure} 

 \begin{figure} 
   \begin{center} 
   \includegraphics[width=1.0\textwidth]{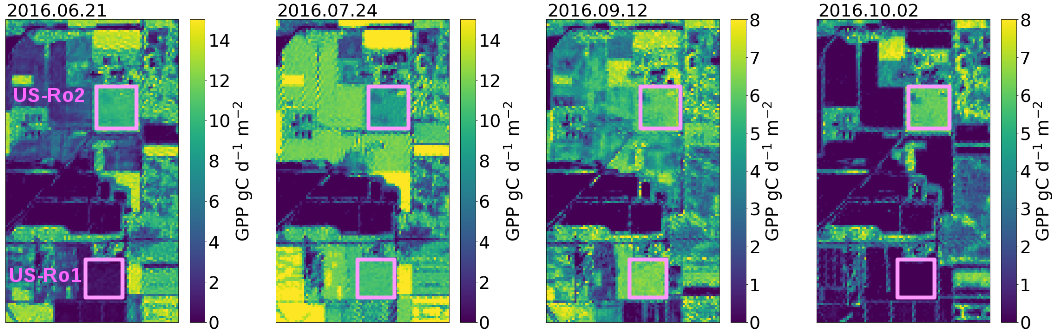}

     \caption{
Maps of estimated GPP {using Sentinel-2 data} over area neighboring the flux tower sites US-Ro1 and   US-Ro2 for four days in summer 2016. 
The fields at these sites are highlighted. Soybeans and Kura Clover were grown at US-Ro1 and US-Ro2 in 2016, respectively.} 
 \label{fig:gpp_retrieval_3} 
   \end{center} 
\end{figure}

The application of our approach to the Sentinel-2 data allows agriculture observations at a sub-field scale. 
The importance of the spatial resolution is very clear when considering areas, which despite being homogeneously croplands, consist in fact of variable crop types. For example, as shown in Figure~\ref{fig:gpp_retrieval_3}, two observed fields (at flux tower sites US-Ro1 and US-Ro2) demonstrate very different phenology. 
Kura Clover (which is a cover crop) was grown in 2016 at the site US-Ro2, and is photosynthetically active during the whole summer. However, the phenology of soybeans is determined by planting and harvesting time, and its growing season is much shorter {than for Kura Clover}. Using Sentinel-2 images allows clear separation of different fields {(not always the case for observations with coarser spatial resolution)}, which improves {the differentiation of GPP among fields and therefore} the estimation of the timing of crop phenology stage.
 
There are generally many issues that can hamper the model performance: the quality of the Sentinel-2 data and their atmospheric correction, the quality and the coarse spatio-temporal resolution of meteorological data, as well as the simplifications and assumptions used within the original model itself {(especially lack of consideration of the soil moisture stress). However, even though there is no direct effect of soil moisture limitation within the model now, a prolonged stress is expected to have an effect on the canopy (e.g., through a reduction in $C_{ab}$, and eventually in LAI) that will be captured later by SCOPE and therefore by our ML model. We note that the applied flux tower dataset covers only the two years when no significant drought stress was affecting crops (even though these fields are not irrigated).
The performance during drought episodes could be improved further by including thermal data \citep{Bayat2018}, which are however not available from Sentinel-2 data.
Besides the limitations of the original model itself, ML algorithms are only a representation of the original model and do not exactly mirror its behavior.
}
We tried to minimize the effects of atmospheric correction by normalizing the reflectance spectra. 
Nevertheless, the model was strongly impacted by the presence of clouds (Figure~\ref{fig:gpp_ro1_rf_nn}).  
{In addition, footprints for the flux towers were not known and therefore we used a simple approach to calculate mean values over the whole fields. The footprints of flux towers for agricultural sites were estimated in previous studies to be up to 1-2 km \citep[e.g.,][]{chen_characterizing_2012, wang_assessment_2016}, but they vary with wind speed and direction, turbulence intensity, surface roughness, measurement height, and atmospheric stability \citep{vesala_flux_2008}, which can also lead to mismatches of flux tower and modeled estimates.}
{Furthermore, we used rather coarse meteorological data (resolution of 0.25\degree) that do not capture finer spatial variability. We also do not explicitly account for the noise in the data, even though the performance of our models is affected by the uncertainties associated with the meteorological and satellite input data as well as the radiative transfer model itself.}
 
Overall, the GPP model performed best for soybeans, for which the relationship between modeled and flux tower data was the most accurate (US-Ro1 and US-Ro5), {and worst for the De-RuS site (Figure~\ref{fig:scatter_gpp}).}
To apply our model to C4 crops, SCOPE simulations need to be redone accounting for the different photosynthetic pathways of the dark reaction of photosynthesis.

\subsection{Application to Landsat 8}
 \label{sec:gpp_landsat8}
{
As the first attempt towards global application of our approach, we tested it on Landsat 8 data. We used the same NN settings as for Sentinel-2, but decreased the number of input satellite bands from ten to six. First, we tested the performance of the model on the Sentinel-2 data at bands shared with Landsat 8, which led to a decrease in model performance ($r^2$=0.77 and RMSE=\SI{2.27}{gC.d^{-1}.m^{-2}}, see Figure~\ref{fig:scatter_gpp_landsat}). The applied model had the same structure as the one chosen for the Sentinel-2 band setting, and therefore it could be expected to perform slightly better if specifically adjusted for Landsat 8.} 
{
Eventually we also applied NN model to the Landsat 8 data in GEE.
The parameters of the model (scaling of the input parameters and the weights and intercepts of the neural network) were exported, and our final NN model was implemented in GEE for Landsat 8 data. The results were tested for the available flux tower data for the four sites in the USA (US-Ro1-2, US-Ro5-6), as the data for the DE-RuS site was only obtained for the Sentinel-2 overpasses. The overall performance of the model was good, but GPP was overestimated outside of the growing season (Figure~\ref{fig:scatter_gpp_landsat}b).}
  
{The better performance of the model for Sentinel-2 bands suggests that the red edge bands do indeed improve ML modeling of GPP, which is also the case for the VI models. These bands have been shown to improve chlorophyll content estimation \citep{clevers_remote_2013}, but in our case they seem to be especially useful for improving the model performance outside of the growing season (both for our ML models as well as for VIs).}

\begin{figure}[h!] 
   \begin{center} 
   \includegraphics[width=0.7\textwidth]{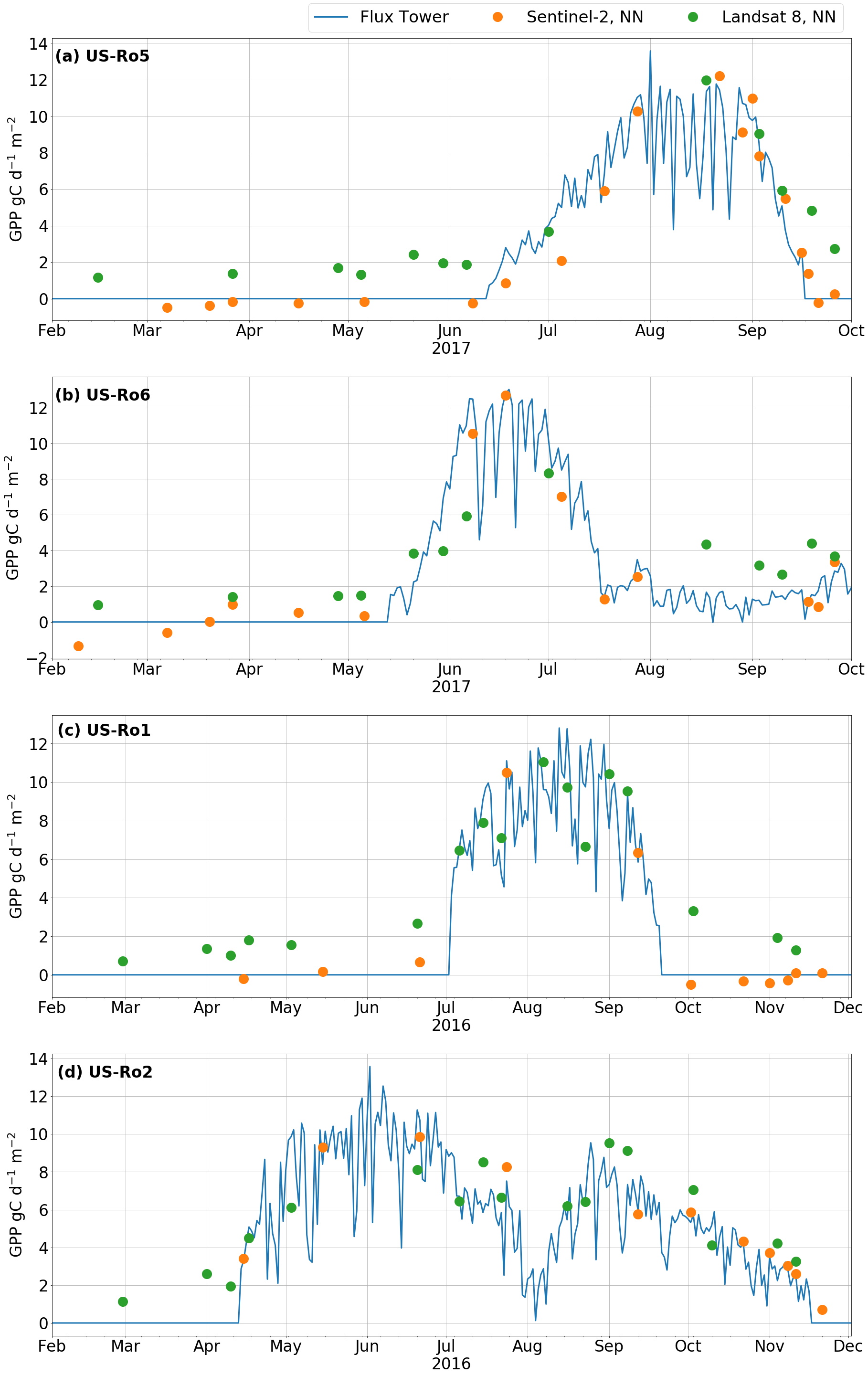}

     \caption{Time series of GPP estimated from flux towers (blue), and modeled with NN using Sentinel-2 (orange) and Landsat 8 (green) data.} 
 \label{fig:gpp_l8_sen2_nn} 
   \end{center} 
\end{figure}

{
Applying our ML models to both instruments simultaneously clearly increases the number of available data points for crop observations (Figure~\ref{fig:gpp_l8_sen2_nn}). Interestingly, for the US-Ro1 site there are many more Landsat 8 observations available as compared to \mbox{Sentinel-2}, despite an overall better revisit time of \mbox{Sentinel-2} (as there were fewer cloudy days during overpasses of Landsat 8). GPP values were quite similar among the sensors, especially during the growing season, which suggests a great potential to extend our approach to other instruments. 
}

\section{Conclusions and Future Work}
\label{sec:summary} 
Estimating photosynthesis of crops is crucial for the crop status monitoring and the forecasting of the agricultural production, and can be greatly supported by satellite remote sensing.
Since recently, and partly due to the advent of Sentinel-2 satellites, an unprecedented amount of data suitable for agriculture observations is available. Taking advantage of recent developments in satellite remote sensing technology, advances in machine learning and more complex and detailed models of photosynthetic processes, we developed a hybrid approach to model GPP with {satellite} data.

We have combined the process-based model SCOPE with ML algorithms to estimate GPP of C3 crops using {satellite} data and ancillary meteorological information.
Several approaches were tested, and our final NN model estimated GPP at the tested flux towers very accurately (with $r^2$ of 0.92 and RMSE of \SI{1.38}{gC.d^{-1}.m^{-2}}). 
ML models were more accurate than VI models, including the reNDVI model {fitted directly} into the flux tower dataset. Our proposed approach successfully estimated GPP across a variety of crop types and environmental conditions, also for time periods of no vegetation. This method was used for high spatio-temporal resolution monitoring of crops with Sentinel-2 {and Landsat 8} data, but can be in fact further extended to other satellites. The results are promising and suggest a way to bridge process-based modeling for global application in an effective manner using a hybrid approach. 
Our model does not use any additional local information from the site, and therefore we plan to apply it globally using platforms providing cloud computing technology. 
{Extending our approach to other sensors, including MODIS, will require additional accounting for spectral differences and the more complex geometry of observations. However, using data covering a longer time span will allow us to use a more extensive flux tower dataset for validation (e.g., FLUXNET2015 dataset), and therefore will provide a good opportunity for further model improvements. This will include testing a selection of the training dataset (e.g., selecting input distribution, assuming dependencies among parameters, adding noise to the data), model types and architectures, as well as procedures performed for harmonization of the datasets among satellites.}
 
\section*{Acknowledgements}
The work by AW, YZ and LG has been funded by the joint project of International Cooperation and Exchange Programs between NSFC and DFG (41761134082) and the Emmy Noether Programme (GlobFluo project) of the German Research Foundation. YZ was also financially supported by the General Program of National Science Foundation of China (41671421). The work by LGC and GMG has been funded by the Spanish Ministry of Economy and Competitiveness (TEC2016-77741-R, ERDF). We thank the PIs and Data Managers of the flux tower sites: Marius Schmidt (Forschungszentrum J\"ulich) and Alexander Graf (Forschungszentrum J\"ulich); John Baker (USDA-ARS), Timothy Griffis (University of Minnesota) and Cody Winker (USDA-ARS).
We acknowledge the European Commission and the Copernicus Open Access Hub for the access to Sentinel-2 data. 
The GLDAS data used in this study were acquired as part of the mission of NASA's Earth Science Division and archived and distributed by the Goddard Earth Sciences (GES) Data and Information Services Center (DISC).
 
\appendix
\section{Appendix}

\subsection{RF}
\label{sec:rf}
In the case of RF model of GPP, we tested settings including maximal depth, minimal samples leaf, as well as changing sample weight (in order to better represent scenarios with small LAI). Similarly to NN, the tests performed on synthetic dataset gave similarly good results, so five models were chosen to be compared with the flux tower data. The performances of these ML algorithms are shown in Table~\ref{table:ML_results_RF}. 

\begin{table}[]
 \scriptsize 
\centering
\caption{Performance ($r^2$) of different ML models on training and test datasets (both with SCOPE), as well as on the validation (val.) data from the flux tower sites. For five different RF models, we varied maximum depth of the tree (MaxD), minimum number of samples required to be at a leaf node (MinLS), as well as sample weights (SW). In the case of the settings SW v.1, we increased the sample weight of data points with GPP below \SI{2}{\micro mol.CO_2.m^{-2}.s^{-1}} to 2, and for the settings SW v.2 to 20.}
\label{table:ML_results_RF}
\resizebox{\textwidth}{!}{%
\begin{tabular}{c*{5}{C}}
  \toprule
  \smash{\raisebox{-.55\normalbaselineskip}{}} & 
        \multicolumn{5}{c}{RF (settings)} \\
  \cmidrule(lr){2-6}	         
	    & \begin{tabular}{@{}c@{}} \#1 \\ default  \end{tabular}   & \begin{tabular}{@{}c@{}} \#2 \\MaxD:20 \end{tabular}   & \begin{tabular}{@{}c@{}} \#3 \\ MinLS:5 \end{tabular}   & \begin{tabular}{@{}c@{}} \#4 \\ SW v.1 \end{tabular}  & \begin{tabular}{@{}c@{}} \#5 \\ SW v.2 \end{tabular}  \\	
        
  \midrule
GPP r$^2$ train &  0.98 & 0.98 & 0.96 & 0.98 & 0.98 \\
GPP r$^2$ test &   0.90 & 0.90 & 0.90 & 0.90 & 0.90 \\
LAI r$^2$ test &  0.51 & 0.51 & 0.52 & 0.51 & 0.51 \\
GPP r$^2$ val. &  0.84 & 0.84 & 0.85 & 0.87 & 0.89 \\
GPP RMSE val. &   1.98 & 1.99 & 1.93 & 1.70 & 1.58 \\
  \bottomrule
\end{tabular}%
}

\end{table}

\begin{figure}[h!] 
   \begin{center} 
   \includegraphics[width=0.78\textwidth]{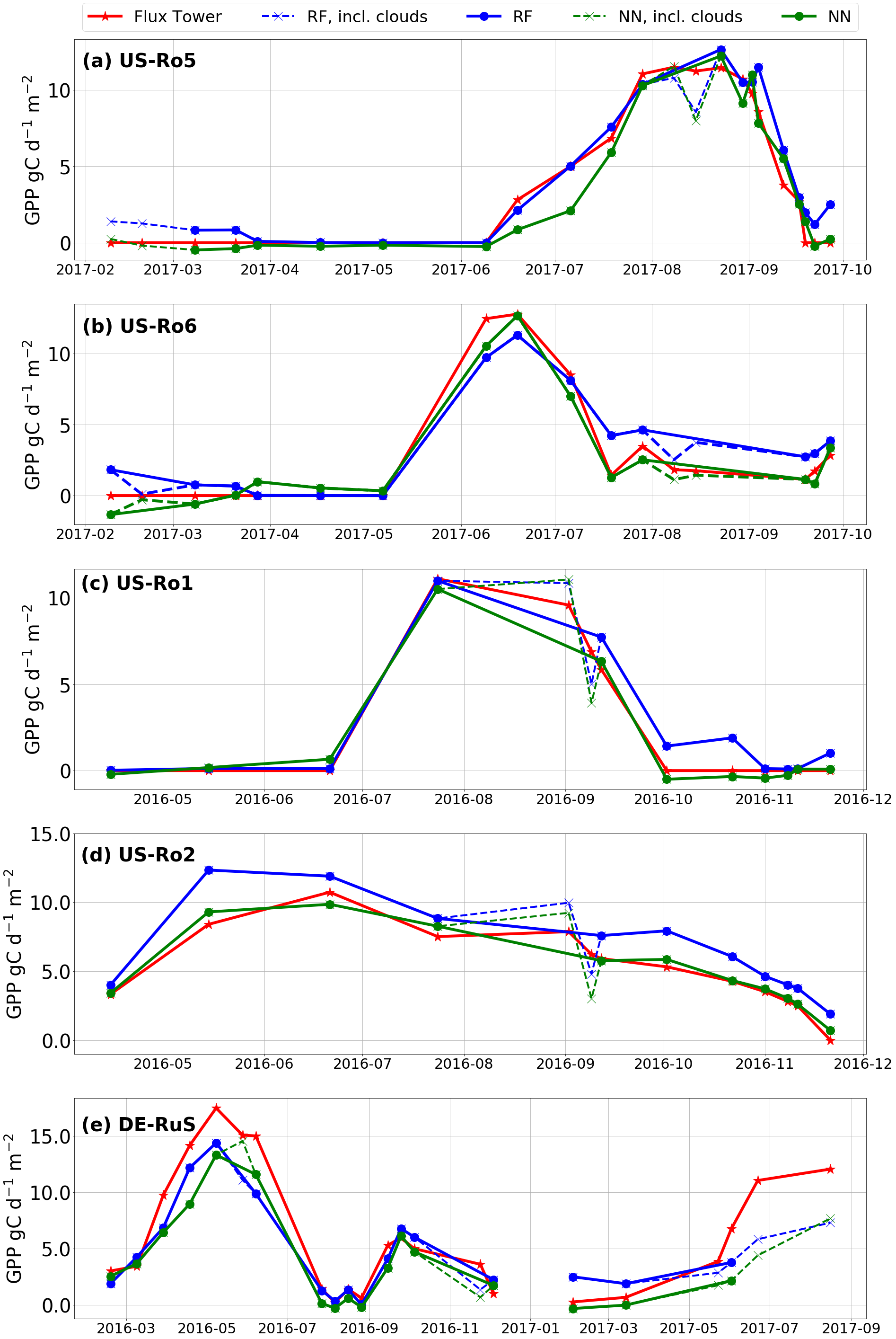}

     \caption{Time series of GPP estimated from flux towers (red), and modeled with NN (green) and RF (blue). The straight line shows the GPP for only clear-sky dates, while dotted images show the dates when fields were covered by thin clouds (that were not removed directly in the visual check of the images).} 
 \label{fig:gpp_ro1_rf_nn} 
   \end{center} 
\end{figure}

\begin{figure} 
   \begin{center} 
   \includegraphics[width=1.0\textwidth]{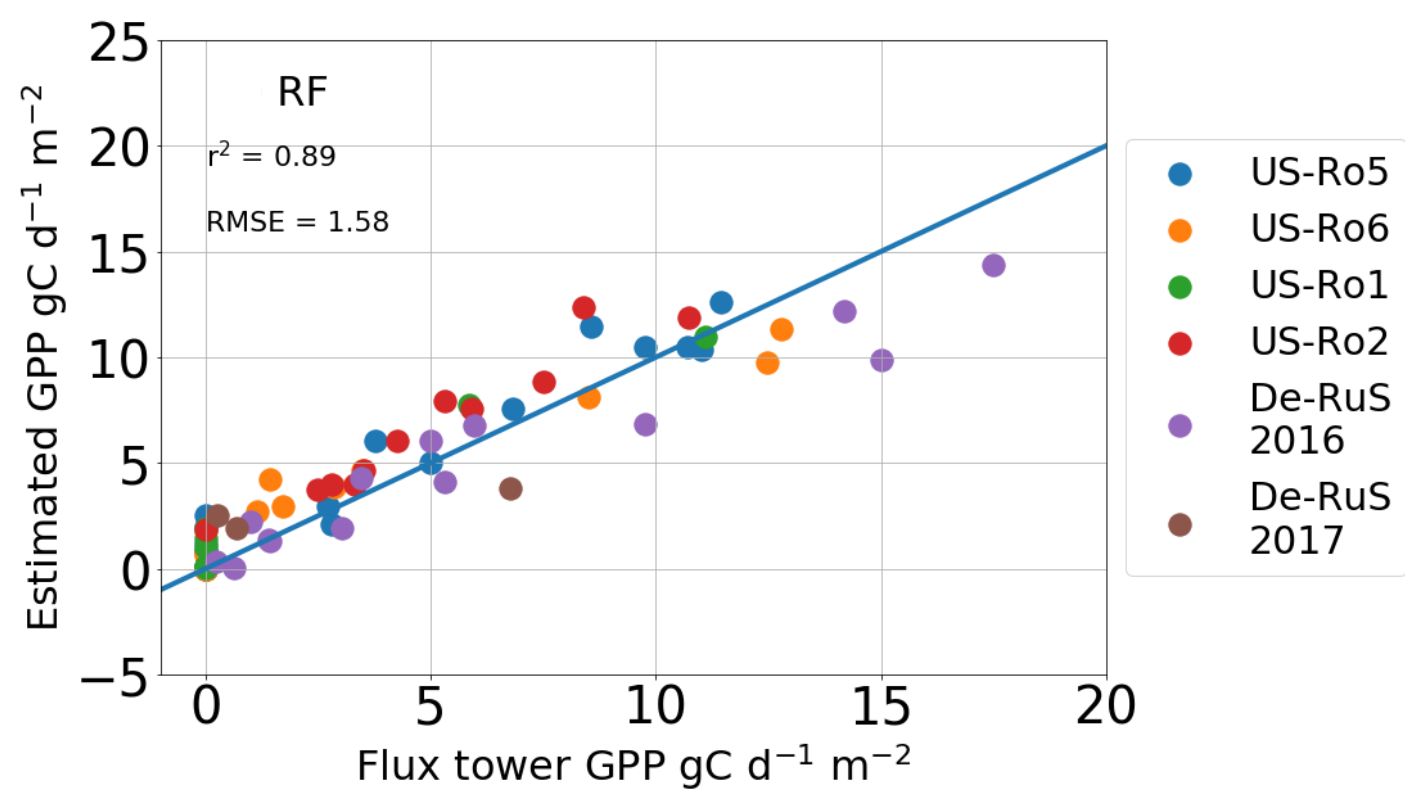}

     \caption{Relationship between daily flux tower GPP and GPP estimated using the RF model. The straight line shows a 1:1 relationship.} 
 \label{fig:scatter_gpp_rf} 
   \end{center} 
\end{figure}

The comparison of resulting time series of GPP measured at flux towers and GPP estimate with RF model (as well as NN model, including scenes covered by thin clouds) are shown in Figure~\ref{fig:gpp_ro1_rf_nn}, and the overall results are compared in scatter plots in Figure~\ref{fig:scatter_gpp_rf}

\end{document}